\documentclass[journal]{IEEEtran}
\IEEEoverridecommandlockouts
\usepackage{cite}
\usepackage{amsmath,amssymb,amsfonts}
\usepackage{graphicx}
\usepackage{textcomp}
\usepackage{xcolor}

\usepackage{microtype}      
\usepackage{graphicx}
\usepackage{tikz}
\usetikzlibrary{shapes,arrows, positioning,arrows.meta,calc}
\usepackage{xfp}
\usepackage{url}            
\usepackage{booktabs}       
\usepackage{amsfonts}       
\usepackage{nicefrac}       
\usepackage{xcolor}         
\usepackage{amsmath}
\usepackage{amssymb} 
\usepackage{amsthm}         
\usepackage{bm}             
\usepackage{algorithm,algpseudocode}
\algblockdefx{MRepeat}{EndRepeat}{\textbf{repeat}}{}
\usepackage{multirow}
\usepackage{array}          
\usepackage[hidelinks]{hyperref}
\usepackage[normalem]{ulem} 
\theoremstyle{plain}

\theoremstyle{definition}

\theoremstyle{remark}

\renewcommand{\hat}{\widehat}
\renewcommand{\bar}{\overline}
\newcommand{\defn}{\triangleq}

\newcommand{\ovec}[1]{\boldsymbol{\Bar{#1}}}
\newcommand{\hvec}[1]{\ensuremath{\Hat{\boldsymbol{#1}}}}
\renewcommand{\vec}[1]{\ensuremath{\boldsymbol{#1}}}

\newcommand{\mat}[1]{\ensuremath{\begin{bmatrix}#1\end{bmatrix}}}

\newcommand{\mc}[1]{\ensuremath{\mathcal{#1}}}

\newcommand{\Real}{{\mathbb{R}}}
\newcommand{\Complex}{{\mathbb{C}}}

\newcommand{\of}[1]{^{\scriptscriptstyle (#1)}}

\newcommand{\tran}{^{\top}}
\newcommand{\herm}{^{\textsf{H}}}
\newcommand*\deriv{\mathop{}\!\mathrm{d}}
\newcommand{\JGibbs}{J}
\newcommand{\true}{_{\mathsf{true}}}
\newcommand{\init}{_{\mathsf{init}}}
\newcommand{\old}{_{\mathsf{old}}}
\newcommand{\new}{_{\mathsf{new}}}
\newcommand{\raw}{_{\mathsf{raw}}}
\newcommand{\sqrtold}[1]{\sqrt{\smash{#1\old}\vphantom{#1}}} 
\newcommand{\sqrtraw}[1]{\sqrt{\smash{#1\raw}\vphantom{#1}}} 
\newcommand{\ssqrt}[1]{\sqrt{\smash{#1}\vphantom{z_r^2}}} 

\DeclareMathOperator{\E}{E}
\DeclareMathOperator{\Cov}{Cov}

\DeclareMathOperator{\tr}{tr}

\DeclareMathOperator{\Diag}{Diag}

\DeclareMathOperator*{\argmin}{arg\,min}

\renewcommand{\eqref}[1]{(\ref{eq:#1})}

\newcommand{\figref}[1]{Fig.~\ref{fig:#1}}
\newcommand{\Figref}[1]{Figure~\ref{fig:#1}}

\newcommand{\Tabref}[1]{Table~\ref{tab:#1}}
\newcommand{\secref}[1]{Sec.~\ref{sec:#1}}
\newcommand{\Secref}[1]{Section~\ref{sec:#1}}
\newcommand{\appref}[1]{App.~\ref{app:#1}}

\renewcommand{\algref}[1]{Alg.~\ref{alg:#1}}
\newcommand{\Algref}[1]{Algorithm~\ref{alg:#1}}
\newcommand{\lineref}[1]{line~\ref{line:#1}}

\newcommand{\textb}[1]{\textcolor{black}{#1}}

\newcommand{\px}{p_{\mathsf{x}}}
\newcommand{\py}{p_{\mathsf{y}}}

\newcommand{\pygx}{p_{\mathsf{y}|\mathsf{x}}}
\newcommand{\pygz}{p_{\mathsf{y}|\mathsf{z}}}

\newcommand{\pzgy}{p_{\mathsf{z}|\mathsf{y}}}
\newcommand{\pxgy}{p_{\mathsf{x}|\mathsf{y}}}
\newcommand{\pyz}{p_{\mathsf{y|z}}}
\newcommand{\pz}{p_{\mathsf{z}}}

\newcommand{\denoiser}{\vec{f}_{\vec{\theta}}}
\newcommand{\diffusion}{\vec{f}_{\mathsf{diff}}}
\newcommand{\dncnn}[1]{\vec{f}_{\mathsf{dncnn\text{-}#1}}}

\begin{document}
\setlength{\arraycolsep}{0.8mm}
\allowdisplaybreaks

\title{Fast and Robust Phase Retrieval via Deep Expectation-Consistent Approximation}
\author{Saurav K. Shastri,~\IEEEmembership{Student Member,~IEEE,} and Philip Schniter,~\IEEEmembership{Fellow,~IEEE}%
\thanks{S. K. Shastri and P. Schniter are with the Dept. of Electrical and Computer Engineering, The Ohio State University, Columbus, OH 43201, USA, Email: \{shastri.19,schniter.1\}@osu.edu.}
}
%

\maketitle

\begin{abstract}
Accurately recovering images from phaseless measurements is a challenging and long-standing problem.
In this work, we present ``deepECpr,'' which combines expectation-consistent (EC) approximation with deep denoising networks to surpass state-of-the-art phase-retrieval methods in both speed and accuracy.
In addition to applying EC in a non-traditional manner, deepECpr includes a novel stochastic damping scheme that is inspired by recent diffusion methods.
Like existing phase-retrieval methods based on plug-and-play priors, regularization by denoising, or diffusion, deepECpr iterates a denoising stage with a measurement-exploitation stage.
But unlike existing methods, deepECpr requires far fewer denoiser calls.
We compare deepECpr to the state-of-the-art prDeep (Metzler et al., 2018), Deep-ITA (Wang et al., 2020), \textb{DOLPH (Shoushtari et al., 2023)}, and Diffusion Posterior Sampling (Chung et al., 2023) methods for noisy phase-retrieval of color, natural, and unnatural grayscale images on oversampled-Fourier and coded-diffraction-pattern measurements and find improvements in both PSNR and SSIM with \textb{significantly} fewer denoiser calls. 
\end{abstract}  


\section{Introduction}
In phase retrieval (PR), one aims to recover an unknown signal or image $\vec{x}\true\in\Real^d$ from phaseless measurements 
\begin{align}
\vec{y} 
= |\vec{Ax}\true| + \vec{w} 
\label{eq:PR_model} ,
\end{align}
where 
$\vec{A}\in\Complex^{m\times d}$ is a known linear operator, 
$|\cdot|$ is an element-wise operation that loses phase information,
and $\vec{w}\in\Real^m$ is noise.
PR is needed when it is infeasible to measure phase information, as often occurs in 
optics \cite{Walther:OA:63}, 
computational biology \cite{Stefik:AI:78}, 
astronomy \cite{Dainty:Chap:87}, 
X-ray crystallography \cite{Millane:JOSAA:90, Harrison:JOSAA:93}, 
coherent diffraction imaging \cite{Miao:Nature:99}, 
speech recognition \cite{Balan:ACHA:06}, 
electron microscopy \cite{Hue:PRB:10},
holography \cite{Zhang:OPT:17}, 
non-line-of-sight imaging \cite{Metzler:OPT:20}, 
and other fields.
PR is challenging because, even with full-rank $\vec{A}$ and no noise, there exist many hypotheses of $\vec{x}$ that explain the phaseless measurements $\vec{y}$ \cite{Shechtman:SPM:15}. 

Various approaches to PR have been proposed.
Classical methods, like the Gerchberg-Saxton (GS) \cite{Gerchberg:OP:72} and Hybrid Input-Output (HIO) \cite{Fienup:AO:82} algorithms, are based on iterative projection. 
Although these algorithms and their variants \cite{Bauschke:JOSAA:03, Elser:JOSAA:03, Chen:PRB:07} are simple to implement and fast to execute, their output qualities don't compete with those of contemporary methods, as we show in the sequel.

A more modern approach is to formulate PR as negative log-likelihood (NLL) minimization and solve it using gradient-based iterative methods.
The PR NLL is non-convex, however, and so spectral initialization strategies have been proposed in an attempt to avoid bad local minima \cite{Candes:TIT:15,Mondelli:COLT:18,Luo:TSP:19}.
Spectral initialization tends to work well with randomized $\vec{A}$ \cite{Candes:TIT:15, Mondelli:COLT:18, Luo:TSP:19} but often fails for the deterministic Fourier $\vec{A}$ (see, e.g., \cite{Metzler:ICML:18}).
As a more direct attack on non-convexity, convex relaxations such as PhaseLift \cite{Candes:CPAM:13} and PhaseMax \cite{Goldstein:TIT:18} have been proposed (see \cite{Fannjiang:AN:20} for an overview). 
However, these relaxations manifest as semidefinite programs in $d^2$-dimensional space, which are computationally impractical for imaging applications with $d$-pixel images.
Approximate Message Passing (AMP) algorithms have also been proposed for PR under sparsity-based signal models \cite{Schniter:TSP:15, Schniter:ASIL:16}.
Although near-optimal for high-dimensional i.i.d.\ or rotationally invariant random $\vec{A}$ \cite{Maillard:NIPS:20, Mondelli:AISTATS:21}, they tend to diverge for deterministic Fourier $\vec{A}$. 

For image PR, various approaches have been proposed to exploit the prior knowledge that $\vec{x}$ is an image. 
For example, the plug-and-play (PnP) \cite{Venkatakrishnan:GSIP:13} and RED \cite{Romano:JIS:17,Reehorst:TCI:19} frameworks have been adapted to PR in \cite{Heide:TG:16, Metzler:ICML:18, Wang:ICML:20b}.
These methods iterate between NLL reduction and denoising, allowing them to harness the power of deep-network-based image denoisers such as DnCNN \cite{Zhang:TIP:17}, and they tend to work well with both random and deterministic $\vec{A}$.
AMP-based PnP PR approaches have also been proposed \cite{Metzler:ICIP:16,Metzler:ICCP:17}, but they tend to struggle with deterministic Fourier $\vec{A}$ \cite{Metzler:ICML:18}.
The Compressed-Sensing Generative Model (GSGM) framework \cite{Bora:ICML:17} was applied to PR in \cite{Hand:NIPS:18}, where---given an image generator $g_{\vec{\theta}}(\cdot)$ with fixed parameters $\vec{\theta}$---one searches for a code vector $\vec{z}$ for which $|\vec{A}g_{\vec{\theta}}(\vec{z})|$ matches the phaseless measurements $\vec{y}$. 
A variation \cite{Wang:Light:20,Bostan:Optica:20} inspired by Deep Image Prior (DIP) \cite{Ulyanov:CVPR:18} fixes $\vec{z}$ and instead optimizes the generator parameters $\vec{\theta}$.
\textb{A further evolution \cite{Chen:TSP:22} applies dropout to $\vec{\theta}$ in an effort to approximate MMSE estimation.
One weakness of these DIP-based approaches is that their inference speed is about 10$\times$ slower than PnP methods like prDeep \cite[Table 4]{Chen:TSP:22}.}
End-to-end deep networks have also been proposed to learn the inverse mapping from $\vec{y}$ to $\vec{x}$ in PR \cite{Sinha:Optica:17, Rivenson:LSA:18}, but they often fail with deterministic $\vec{A}$ \cite{Wang:ICML:20b}.
Recently, diffusion methods have been proposed to sample from the posterior distribution \cite{Torem:ICCV:23,Chung:ICLR:23,Shoushtari:ASIL:23,Xu:24,Wu:24} in PR.
\textb{Some of them \cite{Xu:24,Wu:24} integrate Monte-Carlo sampling to enable tractable theoretical analysis at the expense of slower inference time.
In \secref{experiments}, we test \cite{Chung:ICLR:23,Shoushtari:ASIL:23}, which have similar inference times to PnP methods, but find that they don't perform as well.}

In this work, we propose a novel approach to PR based on the expectation-consistent (EC) approximation framework \cite{Opper:JMLR:05,Fletcher:ISIT:16}, which we refer to as ``deepECpr.''
Different from existing EC-based methods like VAMP \cite{Rangan:TIT:19} and GVAMP \cite{Schniter:ASIL:16}, deepECpr avoids the need for large random $\vec{A}$ by employing EC in a different way.
In addition, deepECpr includes a novel ``stochastic damping'' scheme to further mitigate potential deviations from EC modeling assumptions.
Like PnP, RED, and diffusion-based approaches to PR, deepECpr iteratively calls a deep-net denoiser, but it converges in significantly fewer iterations to more accurate solutions.
We attribute deepECpr's excellent performance to its ability to track error-variances, thereby allowing efficient use of the denoiser. 
This paper builds on our conference publication \cite{Shastri:ASIL:23} by adding stochastic damping, detailed explanations and derivations, and greatly expanded experimental results.

The paper is organized as follows. 
\Secref{background} reviews generalized linear models and EC, 
\secref{method} details our proposed approach, 
\secref{experiments} presents numerical experiments, and 
\secref{conclusion} concludes.

\section{Background}\label{sec:background}

\subsection{Generalized Linear Model (GLM)}\label{sec:GLM}

A wide variety of linear and nonlinear inverse problems fall under the Generalized Linear Model (GLM) framework \cite{McCullagh:Book:89}, which models the relationship between the measurements $\vec{y}\in\mc{Y}^m$ and the signal/image hypothesis $\vec{x}\in\Real^d$ or $\Complex^d$ using 
\begin{align}
\pygx(\vec{y}|\vec{x}) 
= \prod_{i=1}^m \pygz(y_i|z_i) \text{~for~} \vec{z} = \vec{Ax} 
\label{eq:GLM} ,
\end{align}
for scalar ``measurement channel'' $\pygz$ and
forward transform $\vec{A}\in\Real^{m\times d}$ or $\Complex^{m\times d}$. 
By choosing appropriate $\pygz$, one can model, e.g.,
additive noise of an arbitrary distribution,
logistic regression \cite{Hastie:Book:09},
Poisson regression \cite{Figueiredo:TIP:10},
dequantization \cite{Zymnis:SPL:10}, and 
phase retrieval \cite{Shechtman:SPM:15,Dong:SPM:23}. 

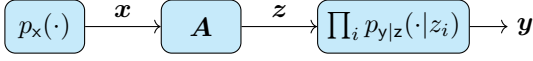
\begin{figure}[t]
  \hspace{8mm}
  \tikzstyle{block} = [draw, rectangle, fill=cyan!20, rounded corners, text centered, minimum width=3em, minimum height=2em]
  \tikzstyle{output} = [text width=3em]
  \begin{tikzpicture}[auto]
  \node[block] (px) {$\px(\cdot)$};
  \node[block, right=1cm of px] (A) {$\vec{A}$};
  \node[block, right=1cm of A] (pygz) {$\prod_i \pygz(\cdot|z_i)$};
  \node[output, right=0.5cm of pygz] (y) {$\vec{y}$};
  \draw [->] (px) -- node[name=x] {$\vec{x}$} (A);
  \draw [->] (A) -- node[name=z] {$\vec{z}$} (pygz);
  \draw [->] (pygz) -- (y);
  \end{tikzpicture}
  \caption{Generalized linear model relating signal $\vec{x}$ to measurements $\vec{y}$.}
  \label{fig:GLM}
\end{figure}

A simple instance of the GLM is the Standard Linear Model (SLM), where $\pygz(y_i|z_i) = \mc{N}(y_i;z_i,v)$ and so
\begin{equation}
\vec{y} = \vec{z}\true + \vec{w} = \vec{Ax}\true + \vec{w}, \quad \vec{w} \sim \mc{N}(\vec{0}, v\vec{I}) 
\label{eq:SLM} .
\end{equation}
Here, $v$ is the variance of the additive white Gaussian noise (AWGN) vector $\vec{w}$.
Through appropriate choice of $\vec{A}$, the SLM covers, e.g.,
denoising, 
deblurring, 
compressed sensing, 
super-resolution, 
inpainting, 
magnetic resonance imaging, 
and
computed tomography. 

For phase retrieval, several variants of $\pygz$ have been employed, such as 
$\pygz(y_i|z_i) = \mc{N}(y_i;|z_i|^2,v)$ \cite{Candes:TIT:15},
$\pygz(y_i|z_i) = \mathrm{Poisson}(y_i;|z_i|^2)$ \cite{Katkovnik:JOSAA:12},
the Rician model resulting from $y_i \sim |z_i+w_i|$ with $w_i\sim\mc{N}(0,v)$ \cite{Schniter:TSP:15}, 
and
\begin{align}
\pygz(y_i|z_i) 
= \mc{N}(y_i;|z_i|,v)
\label{eq:phase_retrieval} .
\end{align}
For algorithm design, we will focus on \eqref{phase_retrieval}, as it has proven to be effective in practice \cite{Yeh:OX:15} and is adopted by state-of-the-art techniques like prDeep \cite{Metzler:ICML:18} and Deep-ITA \cite{Wang:ICML:20b}.

\subsection{Expectation-Consistent Approximation}\label{sec:EC}

We now consider statistical estimation of $\vec{x}\sim \px$ from $\vec{y}$ under the GLM \eqref{GLM}.
When both $\px$ and $\pygz$ are Gaussian, the maximum a posteriori (MAP) and minimum mean-squared error (MMSE) estimates of $\vec{x}$ coincide and are analytically computable.
But Gaussian $\pygz$ is not useful for phase retrieval.
When both $\px$ and $\pygz$ are log-concave, the posterior $\pxgy$ is also log-concave and thus the MAP estimate of $\vec{x}$ can be computed using standard convex-optimization algorithms \cite{Boyd:Book:04,Combettes:Chap:11}. 
Although the $\pygz$ in \eqref{phase_retrieval} is indeed log-concave, high-fidelity image priors $\px$ are not.
Furthermore, the widespread use of PSNR as a recovery-performance metric suggests the use of MMSE estimation over MAP.

The expectation-consistent (EC) approximation framework \cite{Opper:JMLR:05,Fletcher:ISIT:16} is a well established method to approximate the MMSE estimate of $\vec{x}$ from $\vec{y}$, i.e., the conditional mean $\E\{\vec{x}|\vec{y}\}$.
The application of EC to the SLM \eqref{SLM} or GLM \eqref{GLM} is sometimes referred to as vector AMP (VAMP) \cite{Rangan:ISIT:17} and generalized VAMP \cite{Schniter:ASIL:16}, respectively. 
Rigorous analyses of VAMP have established its MSE optimality under high-dimensional rotationally invariant $\vec{A}$ and various assumptions on $\px$ and $\pygz$ \cite{Rangan:TIT:19,Fletcher:NIPS:18,Pandit:JSAIT:20,Takeuchi:TIT:20}.

In the above works, EC is employed to MMSE-estimate $\vec{x}$ from $\vec{y}$ using the prior $\px$ and the likelihood $\pygx$ from \eqref{GLM}.
This approach, however, fails for many deterministic choices of $\vec{A}$,
\textb{%
such as the Fourier-based operators that arise in phase retrieval (see the discussion in \cite{Metzler:ICCP:17,Metzler:ICML:18}) and magnetic resonance imaging (see the discussion in \cite{Millard:OJSP:20,Sarkar:ICASSP:21,Metzler:ICASSP:21,Shastri:JSAIT:22}).
In this paper, we 
}%
instead employ EC to MMSE-estimate $\vec{z}$ from $\vec{y}$, using the prior $\pz$ and likelihood $\pygz$. 
Separately, we estimate $\vec{x}$ from $\vec{z}=\vec{Ax}$.
Thus, when reviewing EC below, we do so in the context of estimating $\vec{z}$ using $\pz$ and $\pygz$.
The more ``traditional'' application of EC (estimating $\vec{x}$ using $\px$ and $\pygx$) is identical up to variable substitutions. 

To begin our review of EC, notice that one can write the true posterior distribution $\pzgy(\cdot|\vec{y})$ without approximation as
\begin{align}
\lefteqn{ \pzgy(\cdot|\vec{y}) 
= \argmin_{q} D\big(q\|\pzgy(\cdot|\vec{y})\big) }\\
&= \argmin_{q} D\big(q\|\pygz(\vec{y}|\cdot)\big) + D(q\|\pz) + H(q) \label{eq:Bayes} \\
&= \argmin_{q_1=q_2=q_3} \underbrace{ D\big(q_1\|\pygz(\vec{y}|\cdot)\big) + D(q_2\|\pz) + H(q_3) }_{\displaystyle \defn \textb{\JGibbs(q_1,q_2,q_3)}}
\label{eq:JGibbs},
\end{align}
where $D(q\|p) \defn\int q(\vec{z})\ln \frac{q(\vec{z})}{p(\vec{z})} \deriv\vec{z}$ denotes Kullback-Leibler (KL) divergence,
$H(q)\defn -\int q(\vec{z}) \ln q(\vec{z}) \deriv\vec{z}$ denotes differential entropy,
and Bayes rule $\pzgy(\vec{z}|\vec{y}) = \pygz(\vec{y}|\vec{z}) \pz(\vec{z}) / \py(\vec{y})$ was used for \eqref{Bayes}.
Because the optimization in \eqref{JGibbs} is intractable, Opper and Winther \cite{Opper:NIPS:05} proposed to relax the equality constraints to moment-matching constraints, i.e.,
\begin{align}
&\argmin_{q_1,q_2,q_3} \textb{\JGibbs(q_1,q_2,q_3)}
\text{~such that~} \label{eq:EC} \\ 
&~ \begin{cases}
	\E\{\vec{z}|q_1\} = \E\{\vec{z}|q_2\} = \E\{\vec{z}|q_3\} \defn \hvec{z} \\
	\tr(\Cov\{\vec{z}|q_1\}) = \tr(\Cov\{\vec{z}|q_2\}) = \tr(\Cov\{\vec{z}|q_3\}) \defn m\hat{v},
   \end{cases} \nonumber
\end{align}
where $\E\{\vec{z}|q_i\}$ and $\Cov\{\vec{z}|q_i\}$ denote the mean and covariance of $\vec{z}$ under $\vec{z}\sim q_i$ and $\tr(\cdot)$ denotes the trace,
and then approximate $\E\{\vec{z}|\vec{y}\}$ by $\hvec{z}$.
The solution to 
\eqref{EC} takes the form
\begin{align}
q_1(\vec{z};\ovec{z}\of{1},\bar{v}\of{1}) 
&\propto \pygz(\vec{y}|\vec{z})\mc{N}(\vec{z};\ovec{z}\of{1},\bar{v}\of{1}\vec{I})~\label{eq:q1}\\
q_2(\vec{z};\ovec{z}\of{2},\bar{v}\of{2}) 
&\propto \pz(\vec{z})\mc{N}(\ovec{z}\of{2};\vec{z},\bar{v}\of{2}\vec{I})~\label{eq:q2}\\
q_3(\vec{z};\hvec{z},\hat{v})
&= \mc{N}(\vec{z};\hvec{z},\hat{v}\vec{I}) 
\label{eq:q3}  
\end{align} 
for some $\ovec{z}\of{1},\ovec{z}\of{2}$ and $\bar{v}\of{1},\bar{v}\of{2}$, known as the ``extrinsic'' means and variances, respectively, \textb{that must be computed (see the discussion after \eqref{pseudolike}), and for}
\begin{align}
\hat{v} &= \big( 1/\bar{v}\of{1} + 1/\bar{v}\of{2} \big)^{-1} 
\label{eq:fp_v}\\
\hvec{z} 
&= \bigg( \frac{\ovec{z}\of{1}}{\bar{v}\of{1}} + \frac{\ovec{z}\of{2}}{\bar{v}\of{2}} \bigg) \hat{v}
\label{eq:fp_z} .
\end{align}
Note that $q_1$ combines the true likelihood $\pygz$ with the Gaussian ``pseudo-prior'' $\vec{z}\sim \mc{N}(\ovec{z}\of{1},\bar{v}\of{1}\vec{I})$, while $q_2$ combines the true prior $\pz$ with the Gaussian ``pseudo-likelihood'' $\mc{N}(\ovec{z}\of{2};\vec{z},\bar{v}\of{2}\vec{I})$, i.e., the model 
\begin{align}
\ovec{z}\of{2}
=\vec{z}\true + \vec{e} \text{~with~} \vec{e}\sim\mc{N}(\vec{0},\bar{v}\of{2}\vec{I})
\label{eq:pseudolike} .
\end{align}

To solve for $(\ovec{z}\of{1},\ovec{z}\of{2},\bar{v}\of{1},\bar{v}\of{2})$, Opper and Winther \cite{Opper:NIPS:05} proposed \algref{EC}, which iterates the following four steps: 
compute the posterior mean and trace-covariance of $q_2(\cdot;\ovec{z}\of{2},\bar{v}\of{2})$, update the extrinsic quantities $(\ovec{z}\of{1},\bar{v}\of{1})$ to obey \eqref{fp_v}-\eqref{fp_z}, compute the posterior mean and trace-covariance of $q_1(\cdot;\ovec{z}\of{1},\bar{v}\of{1})$, and update the extrinsic quantities $(\ovec{z}\of{2},\bar{v}\of{2})$ to obey \eqref{fp_v}-\eqref{fp_z}.
The non-informative initialization of $\bar{v}\of{2}$ and $\ovec{z}\of{2}$ in \lineref{ec_init} ensures that $q_2 = \pz$ initially, so that the initial $\hat{v}\of{2}$ and $\hvec{z}\of{2}$ are the prior variance and mean (as are the initial $\bar{v}\of{1}$ and $\ovec{z}\of{1}$).
\Algref{EC} can be interpreted \cite{Opper:NIPS:05} as an instance of expectation propagation (EP) from \cite{Minka:Diss:01}.

\begin{algorithm}[t]
\caption{EC to infer $\vec{z}\sim\pz$ from $\vec{y}\sim \pygz(\cdot|\vec{z})$}
\label{alg:EC}
\begin{algorithmic}[1]
\Require $\pz(\cdot), \pygz(\vec{y}|\cdot)$  
\State $q_1(\vec{z};\ovec{z}\of{1},\bar{v}\of{1}) \propto \pygz(\vec{y}|\vec{z})\mc{N}(\vec{z};\ovec{z}\of{1},\bar{v}\of{1}\vec{I})$
\State $q_2(\vec{z};\ovec{z}\of{2},\bar{v}\of{2}) \propto \pz(\vec{z})\mc{N}(\vec{z};\ovec{z}\of{2},\bar{v}\of{2}\vec{I})$
\State $\bar{v}\of{2} \gets \infty$ and $\ovec{z}\of{2} \gets \text{arbitrary finite vector in } \Real^m$
\label{line:ec_init}
\MRepeat
    \State // Exploit prior $\pz(\cdot)$ and pseudo-likelihood
    \State
    $\hat{v}\of{2} \gets \tfrac{1}{m}\tr(\Cov\{\vec{z}|q_2(\cdot;\ovec{z}\of{2},\bar{v}\of{2})\})$
    \label{line:ec_vhat2}
    \State
    $\hvec{z}\of{2} \gets \E\{\vec{z}|q_2(\cdot;\ovec{z}\of{2},\bar{v}\of{2})\}$
    \label{line:ec_zhat2}
    \State
    $\bar{v}\of{1} \gets ( 1/\hat{v}\of{2} - 1/\bar{v}\of{2} )^{-1}$  
    \label{line:ec_vbar1}
    \State
    $\ovec{z}\of{1} \gets ( \hvec{z}\of{2}/\hat{v}\of{2} - \ovec{z}\of{2}/\bar{v}\of{2} ) \bar{v}\of{1}$
    \label{line:ec_zbar1}
    \State // Exploit likelihood $\pygz(\vec{y}|\cdot)$ and pseudo-prior
    \State
    $\hat{v}\of{1} \gets \tfrac{1}{m}\tr(\Cov\{\vec{z}|q_1(\cdot;\ovec{z}\of{1},\bar{v}\of{1})\})$
    \label{line:ec_vhat1}
    \State
    $\hvec{z}\of{1} \gets \E\{\vec{z}|q_1(\cdot;\ovec{z}\of{1},\bar{v}\of{1})\}$
    \label{line:ec_zhat1}
    \State
    $\bar{v}\of{2} \gets ( 1/\hat{v}\of{1} - 1/\bar{v}\of{1} )^{-1}$  
    \label{line:ec_vbar2}
    \State
    $\ovec{z}\of{2} \gets ( \hvec{z}\of{1}/\hat{v}\of{1} - \ovec{z}\of{1}/\bar{v}\of{1} ) \bar{v}\of{2}$
    \label{line:ec_zbar2}
\EndRepeat
\Return $\hvec{z}\of{2}$ as the EC approximation of $\E\{\vec{z}|\vec{y}\}$
\end{algorithmic}
\end{algorithm}

\section{Proposed Method} \label{sec:method}

In the following subsections we propose several modifications of the standard EC approach from \algref{EC} that involve
stochastic damping, deep networks,
and simplifications/approximations specific to phase retrieval.

\subsection{Stochastic Damping} \label{sec:damping}

The EC \algref{EC} is not guaranteed to converge for general $\pz$ and $\pygz$.
One technique that helps promote convergence is ``damping'' \cite{Sarkar:TSP:19,Sarkar:ICASSP:21}, which slows the updates in a way that preserves the EC fixed points.
\textb{Damping is also referred to as ``under-relaxation'' in the context of iterative numerical methods and computational fluid dynamics \cite{Ferziger:Book:19}. 
It typically takes the form $\vec{\varphi}\new \gets \mu \vec{\varphi}\raw + (1-\mu) \vec{\varphi}\old $ when iteratively updating $\vec{\varphi}$ using the most recent computation $\vec{\varphi}\raw$.
Here, $\mu\in(0,1]$ is a user-selectable parameter, where lower values increase stability but slow convergence.} 

The damping technique from \cite{Sarkar:ICASSP:21} considers the $\bar{v}\of{i}$ and $\ovec{z}\of{i}$ computed in lines~\ref{line:ec_vbar1}-\ref{line:ec_zbar1} and \ref{line:ec_vbar2}-\ref{line:ec_zbar2} of \algref{EC} as ``raw'' quantities and damps them using the additional steps \textb{(for $i\in\{1,2\}$)}
\begin{subequations} \label{eq:damp}
\begin{align}
\bar{v}\of{i} 
&\gets \big( \mu\of{i} \sqrtraw{\bar{v}\of{i}} + (1-\mu\of{i}) \sqrtold{ \bar{v}\of{i} } \big)^2 
\label{eq:damp_v}\\
\ovec{z}\of{i} 
&\gets \mu\of{i} \ovec{z}\raw\of{i} + (1-\mu\of{i}) \ovec{z}\old\of{i} 
\label{eq:damp_z} ,
\end{align}
\end{subequations}
where $\bar{v}\old\of{i}$ and $\ovec{z}\old\of{i}$ denote the values of $\bar{v}\of{i}$ and $\ovec{z}\of{i}$ from the previous iteration and
$\mu\of{i}\in(0,1]$ is a fixed damping constant.
When $\mu\of{1}=1=\mu\of{2}$, the original EC is recovered.

We now propose a ``stochastic damping'' procedure inspired by diffusion methods like \cite{Ho:NIPS:20}. 
It damps $\bar{v}\of{2}$ as in \eqref{damp_v} but constructs $\ovec{z}\of{2}$ by adding AWGN of a prescribed variance:
\begin{subequations} \label{eq:sdamp}
\begin{align}
\bar{v}\of{2} 
&\gets \big( \mu\of{2} \sqrtraw{\bar{v}\of{2}} + (1-\mu\of{2}) \sqrtold{ \bar{v}\of{2} } \big)^2 
\label{eq:sdamp_v}\\
\ovec{z}\of{2} 
&\gets \ovec{z}\raw\of{2} + \vec{\epsilon}, \quad
\vec{\epsilon}\sim\mc{N}(\vec{0},\textb{\max\{\bar{v}\of{2}-\bar{v}\of{2}\raw,0\}}\vec{I})
\label{eq:sdamp_z} .
\end{align}
\end{subequations}
\textb{To better understand \eqref{sdamp_z}, recall from \eqref{pseudolike} that, under ideal conditions, $\ovec{z}\of{2} = \vec{z}\true + \vec{e}$ for $\vec{e}\sim\mc{N}(\vec{0},\bar{v}\of{2}\vec{I})$ and $\ovec{z}\raw\of{2} = \vec{z}\true + \vec{\varepsilon}$ for $\vec{\varepsilon}\sim\mc{N}(\vec{0},\bar{v}\raw\of{2}\vec{I})$.
This implies that a valid $\ovec{z}\of{2}$ can be constructed from $\ovec{z}\raw\of{2}$ by adding additional AWGN of variance $\bar{v}\of{2}-\bar{v}\raw\of{2}$, as done in \eqref{sdamp_z}.}
Here, it is expected that $\bar{v}\of{2}>\bar{v}\of{2}\raw$ because the variances decrease over the iterations and damping aims to slow down that decrease.
In \secref{experiments}, we show that \eqref{sdamp} has advantages over \eqref{damp}, perhaps because \eqref{sdamp} attempts to enforce the EC model \eqref{pseudolike}.

\subsection{deepEC for general \texorpdfstring{$\vec{A}$}{A}} \label{sec:deepEC}

As formulated in \algref{EC}, EC estimates $\vec{z}\sim\pz$ from observations $\vec{y}\sim \pygz(\cdot|\vec{z})$.
To solve inverse problems using the GLM model \eqref{GLM}, we instead want to estimate $\vec{x}\sim\px$ from $\vec{y}\sim \pygz(\cdot|\vec{z})$ where $\vec{z}=\vec{Ax}$.
In principle, this can be done using a minor modification of \algref{EC} that rewrites $\hvec{z}\of{2}$ in \lineref{ec_zhat2} as a function of an MMSE estimate of $\vec{x}$:
\begin{eqnarray}
\lefteqn{ \hvec{z}\of{2}
= \E\{\vec{z}|q_2(\cdot;\ovec{z}\of{2},\bar{v}\of{2})\} }\\
&=& \E\{\vec{z}|\ovec{z}\of{2}=\vec{z}+\vec{e}\},~\vec{z}\sim\pz,~\vec{e}\sim\mc{N}(0,\bar{v}\of{2}\vec{I}) \\
&=& \vec{A} \underbrace{ \E\{\vec{x}|\ovec{z}\of{2}=\vec{Ax}+\vec{e}\},~\vec{x}\sim\px,~\vec{e}\sim\mc{N}(0,\bar{v}\of{2}\vec{I}) }_{\displaystyle \defn \hvec{x}\of{2}} 
\label{eq:zhat2} ,\qquad
\end{eqnarray}
and then returns $\hvec{x}\of{2}$ as the EC output.
The practical challenge, however, is that computing $\hvec{x}\of{2}$ via \eqref{zhat2} involves MMSE estimation of $\vec{x}$ under the SLM \eqref{SLM}, which is itself non-trivial.

Fortunately, it is now commonplace to train a deep networks to solve SLMs \cite{Arridge:AN:19}.
Writing $\hvec{x}\of{2}\approx \denoiser(\ovec{z}\of{2}; \bar{v}\of{2})$ for deep network $\denoiser$, we could train the network parameters $\vec{\theta}$ via
\begin{align}
\arg\min_{\vec{\theta}} \sum_{t=1}^T \E\big\{\|\vec{x}_t-\denoiser(\vec{Ax}_t+\vec{e}; \bar{v}\of{2})\|^2\big\}
\label{eq:theta}
\end{align}
with training data $\{\vec{x}_t\}_{t=1}^T$, random noise $\vec{e} \sim \mathcal{N}(\vec{0}, \bar{v}\of{2}\vec{I})$, and random noise variance $\bar{v}\of{2} \sim \mathrm{Unif}[0, v_{\max}]$ for some $v_{\max}$.
Likewise, we could train another deep network $h_{\vec{\phi}}$ to approximate the posterior variance $\hat{v}\of{2}$.
Writing $\hat{v}\of{2}\approx h_{\vec{\phi}}(\ovec{z}\of{2}; \bar{v}\of{2})$, its network parameters $\vec{\phi}$ might be trained via
\begin{equation}
\arg\min_{\vec{\phi}} \sum_{t=1}^T \E\big\{\big| \tfrac{1}{m}\|\vec{z}_t-\vec{A}\hvec{x}_t\of{2}\|^2 - h_{\vec{\phi}}(\vec{Ax}_t+\vec{e}; \bar{v}\of{2}) \big|\big\}
\label{eq:phi},
\end{equation}
where $\hvec{x}_t\of{2}\defn \denoiser(\vec{Ax}_t+\vec{e}; \bar{v}\of{2})$ and where $\vec{x}_t$, $\vec{e}$, and $\bar{v}\of{2}$ are constructed as they were for \eqref{theta}.

We will refer to EC with these deep-network approximations as ``deepEC.''
To summarize, deepEC is EC \Algref{EC} but with $\hvec{z}\of{2}$ computed via $\vec{Af}_{\vec{\theta}}(\ovec{z}\of{2}; \bar{v}\of{2})$ and $\hat{v}\of{2}$ computed via $h_{\vec{\phi}}(\ovec{z}\of{2}; \bar{v}\of{2})$.
As such, deepEC solves GLM problems by plugging deep SLM networks into EC \algref{EC}.

\begin{algorithm*}[t]
\caption{deepECpr}
\label{alg:deepECpr}
\begin{algorithmic}[1]
	\Require 
	transform $\vec{A}$,
	channel $\pyz(\vec{y},\cdot)$,
	denoiser $\denoiser(\cdot, \cdot)$, 
	denoising factor $\beta\in(0,1)$,
	image initialization $\hvec{x}\init$, 
	variance initialization $\bar{v}\init$,
	variance initialization factor $\zeta \approx 1.2$,
	damping factors $\mu\of{1}\in(0,1]$ and $\mu\of{2}\in(0,1]$
	\State
	initialize: $\ovec{z}\of{2} \gets \vec{A}\hvec{x}\init + \vec{n}$
	with 
	$\vec{n}\sim\mc{CN}(\vec{0},\bar{v}\init\vec{I})$, and $\bar{v}\of{2} \gets \zeta \bar{v}\init$
	\label{line:init}
	\For{$j=1,\dots,J$}

	\State 
        // Exploit prior $\vec{x}\sim\px$ (via denoiser $\denoiser$) and pseudo-likelihood $p(\ovec{z}\of{2}|\vec{x}) = \mc{CN}(\ovec{z}\of{2};\vec{Ax},\bar{v}\of{2}\vec{I})$
	\State 
        denoise:
	$\hvec{x}\of{2} \gets \denoiser(\Re\{\vec{A}\herm\ovec{z}\of{2}\}, 0.5\bar{v}\of{2})$
	\State 
	deep-network approximated posterior mean and variance:
	$\hvec{z}\of{2} \gets \vec{A}\hvec{x}\of{2}$
	and 
	$\hat{v}\of{2} \gets \beta\bar{v}\of{2}$
	\label{line:DNN}
	\State
	extrinsic variance: 
	$\bar{v}\raw\of{1} \gets ( 1/\hat{v}\of{2} - 1/\bar{v}\of{2} )^{-1}$
        and
        $\bar{v}\old\of{1} \gets \bar{v}\of{1}$
	\label{line:vbar1}
	\State
	extrinsic mean:
	$\ovec{z}\raw\of{1} \gets ( \hvec{z}\of{2}/\hat{v}\of{2} - \ovec{z}\of{2}/\bar{v}\of{2} ) \bar{v}\of{1} $
        and
        $\ovec{z}\old\of{1} \gets \ovec{z}\of{1}$
	\label{line:zbar1}
	\State 
	damping: 
	${\bar{v}\of{1}} \gets \big(\mu\of{1}\sqrt{{\bar{v}\raw\of{1}}} + (1-\mu\of{1})\sqrtold{\bar{v}\of{1}}\big)^2$ and ${\ovec{z}\of{1}} \gets \mu\of{1}{\ovec{z}\raw\of{1}} + (1-\mu\of{1}){\ovec{z}\old\of{1}}$
	\label{line:damp1}

	\State 
        // Exploit likelihood $\pyz(y_i|z_i)$ and pseudo-prior $z_i\sim \mc{CN}(\bar{z}_i\of{1},\bar{v}\of{1})$ for $i=1,\dots,m$ 
	\State
	Laplace-approximated posterior mean and variance: $(\hat{z}_{i}\of{1},\hat{v}_{i}\of{1})~\forall i$ and set $\hat{v}\of{1} \gets \tfrac{1}{m}\sum_{i=1}^{m}\hat{v}_{i}\of{1}$
	\label{line:channel}
	\State
	extrinsic variance: 
	$\bar{v}\raw\of{2} \gets ( 1/\hat{v}\of{1} - 1/\bar{v}\of{1} )^{-1}$
        and
        $\bar{v}\old\of{2} \gets \bar{v}\of{2}$
	\label{line:vbar2}
	\State
	extrinsic mean:
	$\ovec{z}\raw\of{2} \gets ( \hvec{z}\of{1}/\hat{v}\of{1} - \ovec{z}\of{1}/\bar{v}\of{1} ) \bar{v}\of{2} $
	\label{line:zbar2}
	\State
	stochastic damping: $ {\bar{v}\of{2}} \gets \big(\mu\of{2}\sqrtraw{{\bar{v}}\of{2}} + (1-\mu\of{2})\sqrtold{\bar{v}\of{2}}\big)^2$ and ${\ovec{z}\of{2}} \gets \ovec{z}\raw\of{2} + \vec{\epsilon}$ with $\vec{\epsilon} \sim \mc{CN}(\vec{0},[{\bar{v}\of{2}} - \bar{v}\raw\of{2}]\vec{I})$
	\label{line:damp2}
	\EndFor
	\State 
	\Return $\hvec{x}\of{2}$
\end{algorithmic}
\end{algorithm*}

\subsection{deepEC when \texorpdfstring{$\vec{A}\herm\vec{A}=\vec{I}$}{A'*A=I}} \label{sec:deepEC_A}

One practical drawback to deepEC is that the deep networks $\denoiser$ and $h_{\vec{\phi}}$ are dependent on the forward operator $\vec{A}$, and so
the networks must be retrained for each new choice of $\vec{A}$.
There is, however, an important exception to this rule, which is when $\vec{A}\herm\vec{A}=\vec{I}$.
For such column-orthogonal $\vec{A}$, a sufficient statistic \cite{Poor:Book:94} for the estimation of $\hvec{x}\of{2}$ in \eqref{zhat2} is
\begin{align}
\ovec{r}\of{2} \defn \vec{A}\herm\ovec{z}\of{2} 
= \vec{x} + \vec{\epsilon},~~ 
\vec{\epsilon}\sim\mc{N}(\vec{0},\bar{v}\of{2}\vec{I})
\label{eq:suff} ,
\end{align}
in which case $\hvec{x}\of{2}$ solves the MMSE denoising problem
\begin{align}
\hvec{x}\of{2} 
= \E\{\vec{x} | \ovec{r}\of{2} = \vec{x} + \vec{\epsilon}\},~\vec{x}\sim\px,~\vec{\epsilon}\sim\mc{N}(\vec{0},\bar{v}\of{2}\vec{I}) ,
\end{align}
and we can approximate $\hvec{x}\of{2}$ using a deep \emph{denoising} network $\denoiser(\ovec{r}\of{2},\bar{v}\of{2})$ that is invariant to $\vec{A}$.
Likewise, we have that
\begin{align}
\hat{v}\of{2}
&= \tfrac{1}{m}\tr(\Cov\{\vec{z}|\ovec{z}\of{2},\bar{v}\of{2}\}) \\
&= \tfrac{1}{m}\tr(\vec{A}\Cov\{\vec{x}|\ovec{z}\of{2},\bar{v}\of{2}\}\vec{A}\herm) \\
&= \tfrac{1}{m}\tr(\vec{A}\Cov\{\vec{x}|\ovec{r}\of{2},\bar{v}\of{2}\}\vec{A}\herm) \\
&= \tfrac{1}{m}\tr(\vec{A}\herm\vec{A}\Cov\{\vec{x}|\ovec{r}\of{2},\bar{v}\of{2}\}) \\
&= \tfrac{1}{m}\tr(\Cov\{\vec{x}|\ovec{r}\of{2},\bar{v}\of{2}\})
\label{eq:v2} ,
\end{align}
and so the network $h_{\vec{\phi}}$ that deepEC uses to approximate $\hat{v}\of{2}$ can also be invariant to $\vec{A}$.
One way to construct $h_{\vec{\phi}}$ would be to use the Monte-Carlo scheme from \cite{Ramani:TIP:08}, which employs random $\{\vec{p}_c\}_{c=1}^C$ such that $\E\{\vec{p}_c\vec{p}_c\tran\}=\vec{I}$ and small $\delta>0$ in 
\begin{align}
\lefteqn{\tr(\Cov\{\vec{x}|\ovec{r}\of{2},\bar{v}\of{2}\})}\nonumber\\
&\approx \frac{1}{C}\sum_{c=1}^C \frac{\vec{p}_c\tran [\denoiser(\ovec{r}\of{2}\!+\!\delta\vec{p}_c,\bar{v}\of{2}) - \denoiser(\ovec{r}\of{2},\bar{v}\of{2})]}{\delta}
\label{eq:montecarlo}
\end{align}
at the expense of $C\geq 1$ additional calls of the denoiser $\denoiser$.

In summary, for any forward operator $\vec{A}$ that obeys $\vec{A}\herm\vec{A}=\vec{I}$, the proposed deepEC scheme computes approximate-MMSE solutions to GLM problems by iteratively calling a deep denoiser, similar to the PnP \cite{Venkatakrishnan:GSIP:13} or RED \cite{Romano:JIS:17,Reehorst:TCI:19} schemes.
But it requires far fewer iterations, as we demonstrate in \secref{experiments}.

For phase retrieval, $\vec{A}\herm\vec{A}=\vec{I}$ holds for the two classes of $\vec{A}$ that dominate the literature: i) oversampled-Fourier (OSF) and ii) coded diffraction pattern (CDP) \cite{Candes:ACHA:15}.
For OSF, 
\begin{align}
\vec{A} = \vec{F}_m \vec{O} \text{~~for~~} \vec{O}\defn \mat{\vec{I}_d\\ \vec{0}_{m-d\times d}} \in \Real^{m\times d}
\label{eq:OSF},
\end{align}
where $\vec{F}_m\in\Complex^{m\times m}$ is a unitary 2D Fourier transform and $\vec{O}$ pads the vectorized image with zeros.
For CDP,
\begin{align}
\vec{A}=
\frac{1}{\sqrt{K}}
\mat{
\vec{F}_d\Diag(\vec{c}_1)\\[-1mm]
\vdots\\
\vec{F}_d\Diag(\vec{c}_K)
}
\label{eq:CDP} ,
\end{align}
where $\vec{F}_d\in\Complex^{d\times d}$ is a unitary 2D Fourier transform and $\vec{c}_k\in\Complex^d$ are random code vectors with entries drawn independently and uniformly from the unit circle in $\Complex$.
By inspection, $\vec{A}\herm\vec{A}=\vec{I}$ under \eqref{OSF} with $m\geq d$ and \eqref{CDP} with $K\geq 1$.

\subsection{deepEC for Phase Retrieval} \label{sec:deepECpr}

For PR, we propose to apply the deepEC algorithm from \secref{deepEC_A} with the stochastic damping scheme from \secref{damping}.
In addition, we use the minor modifications described below.
The overall scheme, ``deepECpr,'' is detailed in \algref{deepECpr}.

\emph{Initialization:}
Unlike the non-informative $\ovec{z}\of{2}$ initialization used in \algref{EC}, deepECpr uses 
\textb{%
an informative initialization.
For OSF $\vec{A}$, we first run the HIO algorithm \cite{Fienup:AO:82} to produce $\hvec{x}\init$, as in prDeep \cite{Metzler:ICML:18} and Deep-ITA \cite{Wang:ICML:20b}.
For CDP $\vec{A}$, we find that it suffices to run only the first iteration of HIO, and so $\hvec{x}\init = \vec{A}\herm(\vec{y}e^{j\angle{\vec{A}\vec{1}}})$, where $e^{j\angle{\vec{A}\vec{1}}}$ is computed and applied element-wise.
Then, to initialize $\ovec{z}\of{2}$, deepECpr transforms $\hvec{x}\init$}
to the $\vec{z}$-domain and adds circular-complex AWGN:
\begin{equation}
\ovec{z}\of{2} 
\gets \vec{A}\hvec{x}\init + \vec{n}
\text{~with~} \vec{n} \sim \mc{CN}(\vec{0},\bar{v}\init\vec{I}) .
\end{equation}
By using a large noise variance (nominally $\bar{v}\init \approx 100^2$ for pixel amplitudes in $[0,255]$) the structured error artifacts in $\vec{A}\hvec{x}\init$ are suppressed.
Because the error variance of $\ovec{z}\of{2}$ is larger than $\bar{v}\init$ due to the error in $\vec{A}\hvec{x}\init$, the initial $\bar{v}\of{2}$ is set to $\zeta \bar{v}\init$ for some $\zeta>1$ (nominally $\zeta=1.2$).

\emph{Real-valued $\vec{x}$ and complex-valued $\vec{A}$:}
The sufficient statistic $\ovec{r}\of{2}=\vec{A}\herm\ovec{z}\of{2}$ in \eqref{suff} would be appropriate if both $\vec{x}$ and $\vec{A}$ were real-valued, or if both were complex-valued.
However, for the PR problems that we consider, $\vec{x}$ is real-valued and $\vec{A}$ is complex-valued, and so we form the sufficient statistic as 
\begin{align}
\ovec{r}\of{2} \defn \Re\{\vec{A}\herm\ovec{z}\of{2}\} = \vec{x} + \vec{\epsilon}, \quad \vec{\epsilon}\sim\mc{N}(\vec{0},0.5 \bar{v}\of{2}\vec{I})
\label{eq:suff2} ,
\end{align} 
and call the deepECpr denoiser as $\hvec{z}\of{2} = \vec{A}\denoiser(\ovec{r}\of{2},0.5 \bar{v}\of{2})$.

\emph{Simplified variance-prediction:}
As described in \secref{deepEC}, the deep variance predictor $h_{\vec{\phi}}$ aims to estimate the error variance $\hat{v}\of{2}$ of the transformed denoiser output $\hvec{z}\of{2}$.
Although the Monte-Carlo scheme \eqref{montecarlo} is an option, it involves additional denoiser calls, which would increase runtime.
We thus propose to use the simple approximation
\begin{align}
\hat{v}\of{2} \approx \beta \bar{v}\of{2} 
\label{eq:h_simple},
\end{align}
where $\beta\in(0,1)$ is a denoiser-specific constant.
In words, \eqref{h_simple} models the denoiser's output error variance $\hat{v}\of{2}$ as a fixed fraction of its input error variance $\bar{v}\of{2}$.

\emph{Laplace approximation:}
For PR, we adopt the likelihood
\begin{align}
\pygz(y_i|z_i) 
= \mc{N}(y_i;|z_i|,v), 
~~i=1,\dots,m
\label{eq:phase_retrieval2} ,
\end{align}
due to its excellent performance in practice \cite{Yeh:OX:15}.
However, the resulting EC posterior 
\begin{align}
q_2(z_i;\bar{z}\of{1}_i,\bar{v}\of{1}) 
&= \frac{\pygz(y_i|z_i) \mc{CN}(z_i;\bar{z}\of{1}_i,\bar{v}\of{1})} 
       {\int \pygz(y_i|z_i') \mc{CN}(z_i';\bar{z}\of{1}_i,\bar{v}\of{1}) \deriv z_i'} \\
&\defn \pzgy(z_i|y_i;\bar{z}\of{1}_i,\bar{v}\of{1})
\end{align}
does not have analytically tractable mean or variance.
Consequently, we employ the Laplace approximation \cite{Bishop:Book:07}, which assigns $\hat{z}_i\of{1}$ to the posterior mode (i.e., the MAP estimate) and $\hat{v}_i\of{1}$ to the trace of the inverse Hessian of $-\ln \pzgy(z_i|y_i;\bar{z}\of{1}_i,\bar{v}\of{1})$ at $z_i=\hat{z}_i\of{1}$, both of which have closed-form expressions.
In particular, the MAP estimate of $z_i$ is given by \cite{Wang:ICML:20b}
\begin{align}
\hat{z}\of{1}_i = \frac{\bar{v}\of{1}y_i + 2v|\bar{z}\of{1}_i|}{\bar{v}\of{1} + 2v} e^{j \angle \bar{z}\of{1}_i}
\label{eq:MAP} ,
\end{align}
and the trace-inverse-Hessian at $z_i=\hat{z}_i\of{1}$ is
\begin{align}
	\hat{v}_i\of{1} = \frac{\bar{v}\of{1}(\bar{v}\of{1}y_i + 4v|\bar{z}\of{1}_i|)}{2|\bar{z}\of{1}_i|(\bar{v}\of{1} + 2v)} 
	\label{eq:hat_var_of_1_i} 
\end{align}
as derived in \appref{hess}.
The overall posterior variance $\hat{v}\of{1}$ is then set to the average value over all $i$, i.e., $\hat{v}\of{1} = \tfrac{1}{m}\sum_{i=1}^{m}\hat{v}_{i}\of{1}$.


\begin{figure*}[h]
	\centering
	\includegraphics[width=\linewidth,trim=5 5 5 5,clip]{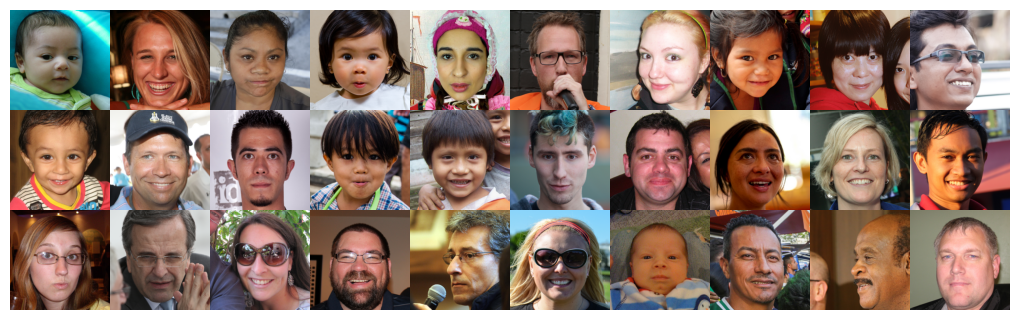}
	\caption{The thirty $256\times 256$ FFHQ test images.}
	\label{fig:FFHQ_test}
\end{figure*}

\begin{figure}[t]
\includegraphics[width=0.48\linewidth,trim=7 9 5 6,clip]{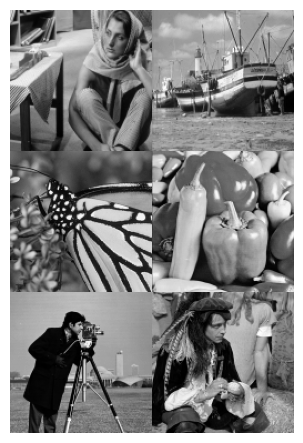}
\hfill
\includegraphics[width=0.48\linewidth,trim=7 9 5 6,clip]{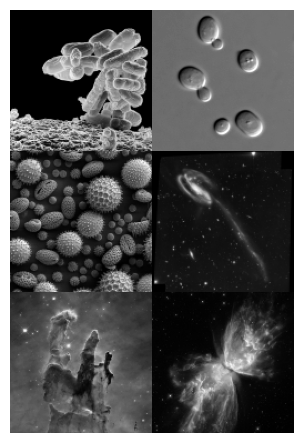}
\caption{The natural (left) and unnatural (right) grayscale test images.}
\label{fig:grayscale_test}
\end{figure}

\section{Numerical Experiments}\label{sec:experiments}

In this section, we present numerical experiments with Poisson-shot-noise-corrupted OSF and CDP phaseless measurements, comparing the performance of the proposed deepECpr algorithm to the classical HIO \cite{Fienup:AO:82} and state-of-the-art prDeep \cite{Metzler:ICML:18}, Deep-ITA (both ``F'' and ``S'' variants) \cite{Wang:ICML:20b}, \textb{Diffusion Posterior Sampling (DPS) \cite{Chung:ICLR:23}, and Diffusion mOdeL for PHase retrieval (DOLPH) \cite{Shoushtari:ASIL:23}} approaches.
Our code is available at \texttt{\url{https://github.com/Saurav-K-Shastri/deepECpr}}.

\subsection{Test Data and Denoisers}

We test on three image datasets.
As in the DPS paper \cite{Chung:ICLR:23}, we use $256 \times 256$ color images from the FFHQ test partition \cite{Karras:CVPR:19}. 
In particular, we use the first thirty images, shown in \figref{FFHQ_test}.
As in the Deep-ITA paper \cite{Wang:ICML:20b}, we use the six ``natural'' $128 \times 128$ grayscale images shown in \figref{grayscale_test},
and as in the prDeep paper \cite{Metzler:ICML:18}, we use the six ``unnatural'' $128 \times 128$ grayscale images shown in \figref{grayscale_test}. 

\newcommand{\noisy}{_{\mathsf{noisy}}}
\newcommand{\inp}{_{\mathsf{in}}}
When recovering the FFHQ test images, we use a denoiser that was constructed by rescaling (see \cite[eq.\ (9)]{Chung:ICLR:23}) an unconditional DDPM diffusion model \cite{Ho:NIPS:20} trained on 49\,000 FFHQ training images \cite{Chung:github:23}.
The resulting diffusion denoiser ``$\diffusion(\vec{r}\noisy,v\inp)$'' accepts a noisy image $\vec{r}\noisy$ and an input-noise variance $v\inp$, and thus is directly compatible with deepECpr, \textb{DOLPH}, and DPS. 
The prDeep and Deep-ITA methods, however, do not have a way to estimate $v\inp$ and instead sequence through a bank of four ``blind'' denoisers designed to work at input-noise standard deviations (SD) of 60, 40, 20, 10 (relative to pixel values in $[0,255]$), respectively.
Thus, for prDeep and Deep-ITA, we use $\diffusion(\vec{r}\noisy,v\inp)$ with $v\inp\in\{60^2, 40^2, 20^2, 10^2\}$ for FFHQ image recovery.

When recovering the grayscale images, we follow prDeep and Deep-ITA in using a bank of four blind DnCNN \cite{Zhang:TIP:17} denoisers $\{\dncnn{60}(\cdot), \dncnn{40}(\cdot), \dncnn{20}(\cdot), \dncnn{10}(\cdot)\}$.
Although deepECpr would likely work better with a non-blind denoiser, we wanted a fair comparison with prDeep and Deep-ITA. 
These DnCNN denoisers were trained over the SD intervals $\{[40,60],[20,40],[10,20],[0,10]\}$, respectively, on the BSD400 \cite{bsd400} dataset using the bias-free approach from \cite{Mohan:ICLR:20}.
Note that our natural grayscale test images are similar to (but distinct from) the BSD400 training data, while our unnatural grayscale test images could be considered out-of-distribution \textb{for this denoiser.
For DOLPH and DPS, which are incompatible with blind denoisers, we use a pretrained ImageNet diffusion denoiser \cite{Dhariwal:NIPS:21} and copied the grayscale measurements into all three color channels.}

\subsection{Measurement Generation}

Like the prDeep and Deep-ITA papers, we employ OSF and CDP forward operators $\vec{A}$ (recall \eqref{OSF}-\eqref{CDP}) with 4$\times$ oversampling and the \textb{(Gaussian approximated)} Poisson shot-noise mechanism from \cite{Metzler:ICML:18}
\begin{align}
y_i^2 = |z_i|^2+w_i \text{~with~} w_i\sim \mc{N}(0,\alpha^2 |z_i|^2)
\label{eq:noise_model} ,
\end{align}
where $y_i^2/\alpha^2\sim\mc{N}(|z_i|^2/\alpha^2,|z_i|^2/\alpha^2)$ is approximately $\mathrm{Poisson}((|z_i|/\alpha)^2)$ for sufficiently small noise levels $\alpha$.
\textb{Note that, because we scale $\vec{A}$ to have unit-norm columns, our $\alpha$ may have a different meaning than the $\alpha$ used in other papers.
For example, with OSF measurements, our $\alpha$ must be scaled by $1/\sqrt{K}=0.5$ to match the prDeep paper \cite{Metzler:ICML:18} 
or by $\sqrt{K}=2$ to match the Deep-ITA paper \cite{Wang:ICML:20b}.
With CDP measurements, our $\alpha$ must be scaled by $\sqrt{K}=2$ to match the prDeep paper.}

\subsection{Performance Evaluation}

We quantify performance using PSNR and SSIM \cite{Wang:TIP:04} after resolving certain ambiguities fundamental to PR. 

Phaseless Fourier measurements are unaffected by spatial translations, conjugate flips, and global phase \cite{Bendory:SB:15}. 
The phase ambiguity is circumvented when the image pixels are non-negative, as with all of our test images.
Furthermore, the translation ambiguity is avoided when oversampling is used with images that have a sufficient number of non-zero edge pixels, as with our color FFHQ test images, but not our grayscale test images.
Thus, when recovering the grayscale test images from OSF measurements, we correct for both translations and flips, but when recovering the color FFHQ test images, we correct for flips only. 
With color images, however, one further complication arises with HIO. 
Because HIO separately recovers each color channel, the flip ambiguity can manifest differently in each channel.
To address this latter problem, we fix the first channel of the HIO recovery and flip of each remaining channel to maximize its correlation with the first channel.

Phaseless CDP measurements experience only a global-phase ambiguity \cite{Fannjiang:IP:12}, which is inconsequential with our \textb{non-negative} test images, and so we perform no ambiguity resolution.

\subsection{Algorithm Setup} \label{sec:algs}

For HIO, we ported the MATLAB implementation from \cite{Metzler:github:18} to Python and set the tunable step-size parameter to $0.9$.
For prDeep, we use the Python implementation from \cite{Hekstra:github:18}.
For Deep-ITA \textb{and DOLPH}, we implemented the algorithm in Python/PyTorch since no public information is available.
For DPS, we use the Python implementation from \cite{Chung:github:23}.
As in the prDeep and Deep-ITA papers, we use $\pygz$ from \eqref{phase_retrieval} for all algorithms (except HIO, which does not use a likelihood).
For deepECpr, we use damping factors $\mu\of{1} = 0.3$ and $\mu\of{2} = 0.075$ and initialization factor $\zeta = 1.2$.
EM-based auto-tuning \cite{Fletcher:NIPS:17} is used to re-evaluate $\bar{v}\of{2}$ for the first 3 and 10 iterations in the FFHQ and grayscale experiments, respectively.
The initial noise level $\bar{v}\init$ is set to $120^2$ for the FFHQ experiments and $70^2$ for the grayscale experiments.
Experiment-specific settings are described later. 

\emph{Initialization/Reporting for OSF:}
For the OSF experiments, we apply the initialization/reporting protocol used in both the prDeep and Deep-ITA papers to those algorithms, as well as to HIO and deepECpr.  
First, HIO is run 50 times, for 50 iterations each, from a random initialization. 
The candidate $\hvec{x}$ with the lowest measurement residual $\|\vec{y}-|\vec{Ax}|\|$ is then selected as the HIO initialization, after which HIO is run for 1000 iterations.
In the case of color images, the second and third channels are flipped to best match the first. 
Finally, the resulting HIO output is used to initialize prDeep, Deep-ITA, and deepECpr.
The entire procedure is repeated three times, and the reconstruction with the lowest measurement residual is reported as the final output for each algorithm.
For DPS \textb{and DOLPH}, we follow the  \textb{DPS} authors' recommendation to run the algorithm four times and report the reconstruction with the lowest measurement residual.

\emph{Initialization/Reporting for CDP:}
For the CDP experiments, we use the initialization 
$\hvec{x}\init = \vec{A}\herm(\vec{y}e^{j\angle{\vec{A}\vec{1}}})$ 
for HIO, prDeep, Deep-ITA, and deepECpr, where $e^{j\angle{\vec{A}\vec{1}}}$ is computed and applied element-wise.
Each algorithm was run only once.

\subsection{OSF Phase Retrieval of FFHQ Images} \label{sec:FFHQ_OSF}

\begin{table*}[t]
	\centering
	\caption{Average PSNR and SSIM for FFHQ phase retrieval with OSF and CDP operators and noise level $\alpha$.}
	\resizebox{\linewidth}{!}{
		\begin{tabular}{|c|c|c|c|c|c|c|c|c|c|c|c|c|}
			\hline
			\multirow{3}{*}{method }& \multicolumn{6}{c|}{OSF}& \multicolumn{6}{c|}{CDP}\\
						\cline{2-13}
			& \multicolumn{2}{c|}{$\alpha = 4$} & \multicolumn{2}{c|}{$\alpha = 6$} & \multicolumn{2}{c|}{$\alpha = 8$}& \multicolumn{2}{c|}{$\alpha = 5$} & \multicolumn{2}{c|}{$\alpha = 15$} & \multicolumn{2}{c|}{$\alpha = 45$}\\
						\cline{2-13}
			& PSNR & SSIM  & PSNR & SSIM & PSNR & SSIM & PSNR & SSIM  & PSNR & SSIM & PSNR & SSIM \\
			\hline
			HIO  & 27.37   & 0.6759 &  26.08 & 0.6163 & 25.03  & 0.5664 & 36.48  & 0.9140 &  26.79 & 0.6064 & 17.58  &  0.2330\\ 
			Deep-ITA-F & 35.05  & 0.9420  & 34.94  & 0.9374 &  34.50 & 0.9321 & 41.69  & 0.9786 & 37.20  & 0.9556 &  29.93 & 0.8402\\ 
			Deep-ITA-S & 34.01  & 0.9365  &  34.54 & 0.9367 & \textbf{35.15 } & \textbf{0.9412} & 41.87  & 0.9795 & 37.15  & 0.9531 & 27.28  & 0.7817\\ 
			prDeep & \underline{37.69}  & \underline{0.9654}  &  \underline{35.28} & \underline{0.9523} & 33.68  & \underline{0.9410} &  \underline{42.43} & \underline{0.9816} & \underline{37.48 } & \underline{0.9584} &  23.14 & 0.4477\\ 
			\textb{DOLPH}  & \textb{14.65} & \textb{0.3426} & \textb{14.65} & \textb{0.3420} & \textb{14.64} & \textb{0.3393} & \textb{40.74} & \textb{0.9755} &  \textb{33.84} & \textb{0.8806} & \textb{20.98} &  \textb{0.3541} \\
			DPS & 27.22  & 0.7674 &  25.84 & 0.7530 &  24.57 & 0.7408 & 41.52 & 0.9786  &  35.68 &  0.9381 & \underline{30.12} & \underline{0.8428} \\ 
			deepECpr (proposed) & \textbf{39.75}   & \textbf{0.9720}  & \textbf{37.01} &  \textbf{0.9567} & \underline{34.86} & 0.9404 & \textbf{43.12}  & \textbf{0.9846}  & \textbf{37.55}  & \textbf{0.9589} & \textbf{32.15}  & \textbf{0.8941} \\ 
			\hline
	\end{tabular}}
	\label{tab:FFHQ_OSF_CDP}
\end{table*}

\begin{figure*}[t]
	\centering
	\includegraphics[width=\linewidth,trim=5 5 5 5,clip]{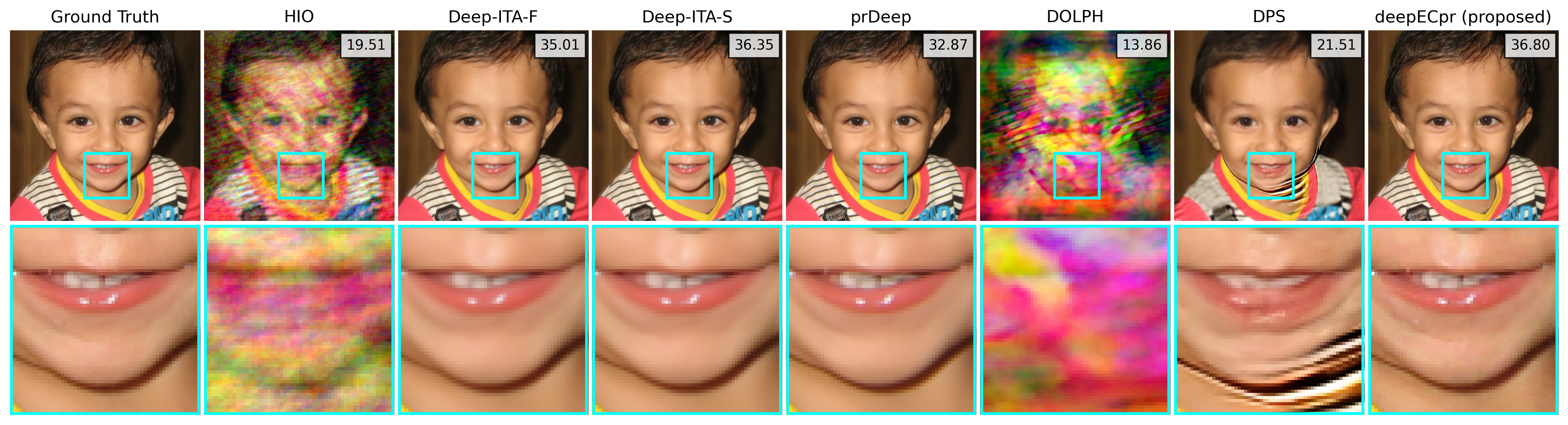}
	\vspace{-5mm}
	\caption{\textb{Top: Example FFHQ image recoveries from phaseless OSF measurements at noise level $\alpha=8$, with PSNR indicated in the top right corner of each image. Bottom: Zoomed versions of the cyan regions in the top row. Note that the HIO, DOLPH, and DPS recoveries contain strong artifacts and that Deep-ITA and prDeep show oversmoothing.}}
	\label{fig:FFHQ_OSF_recon1}
\end{figure*}


\begin{figure}[t]
	\centering
	\includegraphics[width=0.9\linewidth,trim=7 8 5 6,clip]{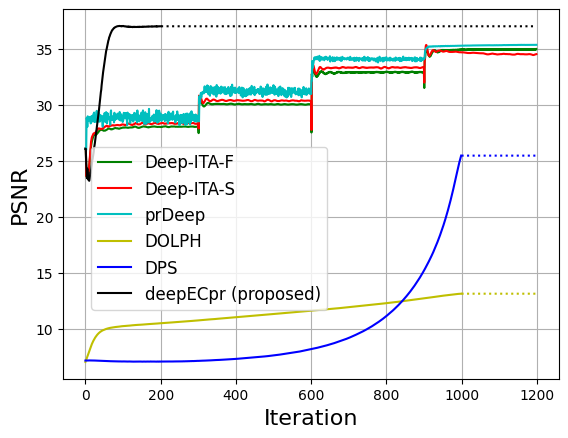}
	\caption{\textb{Average PSNR versus iteration for FFHQ phase retrieval from OSF measurements at noise level $\alpha=6$.}}
	\label{fig:PSNRvsIter_OSF_FFHQ}
\end{figure}

\begin{figure}[t]
	\centering
	\includegraphics[width=0.9\linewidth,trim=4 7 5 6,clip]{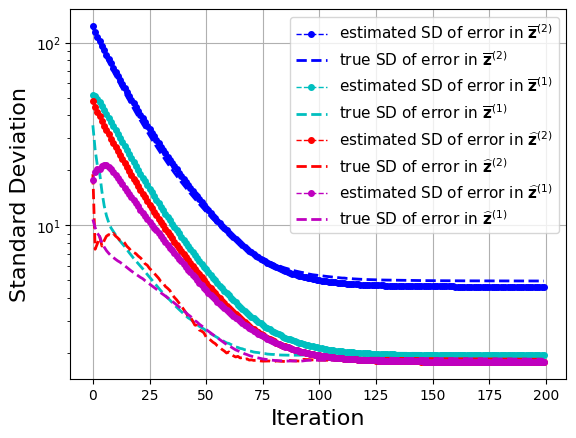}
	\caption{Evolution of the true and estimated $\vec{z}$ errors in deepECpr for OSF phase retrieval at noise level $\alpha = 8$.}
	\label{fig:SD_evolution_example}
\end{figure}

We first investigate the recovery of FFHQ test images from noisy, phaseless OSF measurements.
For prDeep and Deep-ITA, we sequence through the denoisers $\diffusion(\cdot,60^2), \diffusion(\cdot,40^2), \diffusion(\cdot,20^2)$, and $\diffusion(\cdot,10^2)$ for $300$ iterations each, for a total of $1200$ iterations, as recommended by the authors.
Also, we set the $\lambda$ parameter to $0.1\sqrt{v}$ for prDeep, and $0.05\sqrt{v}$ for Deep-ITA, to maximize validation PSNR.
For DPS, we use $1000$ steps as recommended by the authors and set the scale parameter to $0.0075$ to maximize validation PSNR.
\textb{Similarly, we use $1000$ steps for DOLPH and set the step-size parameter to $10^{-6}$ to maximize validation PSNR.}
For deepECpr, we use $\diffusion$ as specified in \algref{deepECpr} for $200$ iterations with denoising factor $\beta=0.15$.

\Tabref{FFHQ_OSF_CDP} shows the average PSNR and SSIM at noise levels $\alpha \in\{ 4, 6, 8\}$. 
There, deepECpr outperforms the other methods in both PSNR and SSIM at $\alpha \in\{ 4, 6\}$, where the PSNR advantage is significant ($\approx 2$~dB).
At $\alpha = 8$, deepECpr trails the best-performing technique, Deep-ITA-S, by a small margin: $0.29$~dB 
in PSNR and $0.0008$ 
in SSIM.

\Figref{FFHQ_OSF_recon1} shows example recoveries at $\alpha = 8$.
There we see severe artifacts in the HIO, \textb{DOLPH}, and DPS recoveries.
Although the prDeep, Deep-ITA-F, Deep-ITA-S, and deepECpr approaches avoid unwanted artifacts, the proposed deepECpr does the best job of recovering fine details in the ground-truth image.

\Figref{PSNRvsIter_OSF_FFHQ} plots PSNR (averaged over the FFHQ test images) versus iteration at $\alpha = 6$. 
Note that the prDeep, Deep-ITA, \textb{DOLPH}, DPS, and deepECpr algorithms all call the denoiser once per iteration and thus have similar per-iteration complexities.
From the figure, we see that deepECpr converges \textb{significantly} faster than its competitors.

\Figref{SD_evolution_example} plots the standard-deviations (SDs) of the deepECpr-estimated and true errors in the $\vec{z}$ estimates for an typical OSF recovery at noise level $\alpha = 8$.
In particular, it plots $\sqrt{\bar{v}\of{i}}$ in comparison to the true error $m^{-1/2}\|\ovec{z}\of{i}-\vec{z}\|$, and $\sqrt{\hat{v}\of{i}}$ in comparison to the true error $m^{-1/2}\|\hvec{z}\of{i}-\vec{z}\|$, for $i\in\{1,2\}$.
The figure shows the accuracy of $\bar{v}\of{2}$ over all iterations and the accuracy of the other estimates after iteration $100$.
Furthermore, it shows the EC fixed-point condition $\hat{v}\of{1}=\hat{v}\of{2}$ being satisfied. 

\begin{figure}[t]
	\centering
	\includegraphics[width=0.9\linewidth,trim=4 7 5 6,clip]{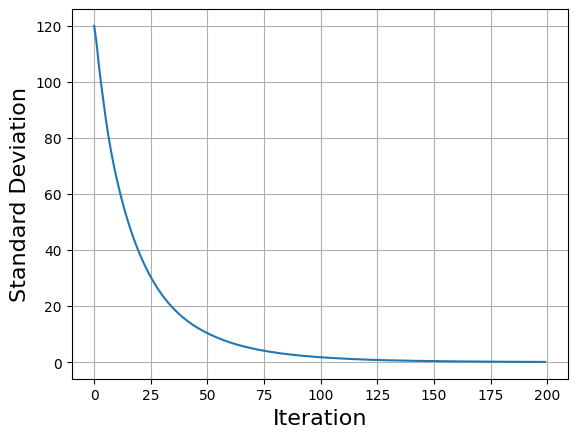}
	\caption{Standard deviation of the noise added by deepECpr's stochastic damping scheme versus iteration for an example of FFHQ phase retrieval with OSF measurements at noise level $\alpha = 8$.}
	\label{fig:Added_SD_evolution_example}
\end{figure}

Table~\ref{tab:FFHQ_OSF_damping_ablation} presents an ablation study comparing the proposed stochastic damping \eqref{sdamp} to the deterministic damping \eqref{damp} when used in \lineref{damp2} of \algref{deepECpr}. 
With stochastic damping, deepECpr yields higher PSNR and SSIM values at all tested noise levels $\alpha$.
\Figref{Added_SD_evolution_example} plots the SD of the AWGN noise injected by stochastic damping as a function of the deepECpr iteration.
The plot shows that the SD starts high but decreases to zero with the iterations.

\subsection{CDP Phase Retrieval of FFHQ Images} \label{sec:FFHQ_CDP}

\begin{figure*}[t]
	\centering
	\includegraphics[width=\linewidth,trim=5 5 5 5,clip]{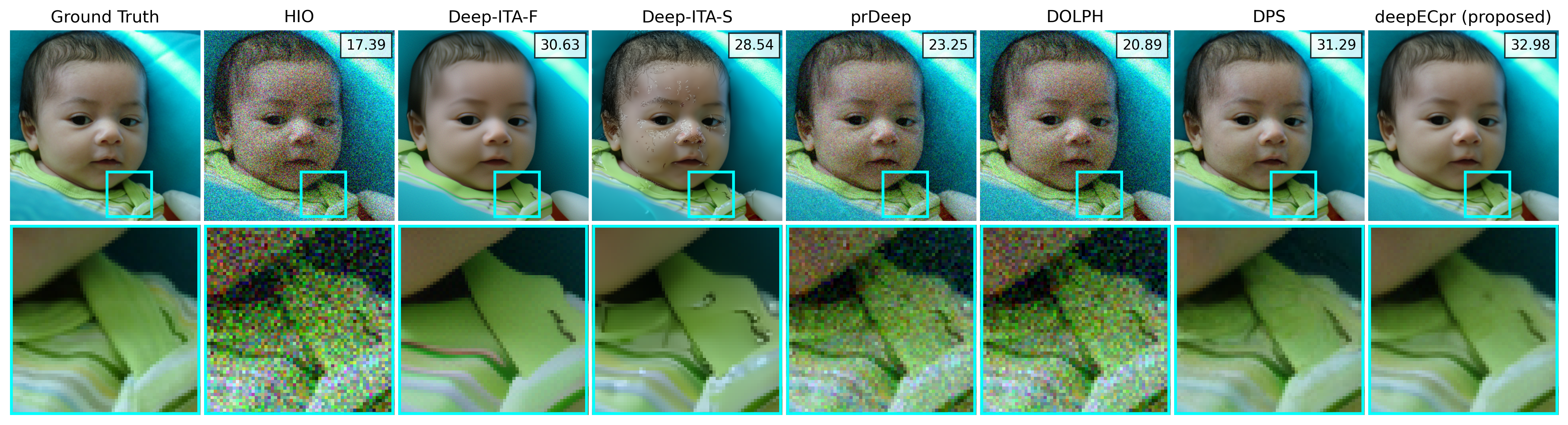}
	\vspace{-5mm}
	\caption{\textb{Top: Example FFHQ image recoveries from phaseless CDP measurements at noise level $\alpha=45$, with PSNR indicated in the top right corner of each image. Bottom: Zoomed versions of the cyan regions in the top row. Note that the HIO, prDeep, DOLPH recoveries are very noisy and that Deep-ITA is plagued by both oversmoothing and hallucinations.}}
	\label{fig:FFHQ_CDP_recon1}
\end{figure*}

\begin{table}[t]
	\centering
	\caption{PSNR and SSIM of deterministic versus stochastic damping for FFHQ OSF phase retrieval under noise level $\alpha$}
	\resizebox{\columnwidth}{!}{
		\begin{tabular}{|c|c|c|c|c|c|c|}
			\hline
			type of damping used & \multicolumn{2}{c|}{$\alpha = 4$} & \multicolumn{2}{c|}{$\alpha = 6$} & \multicolumn{2}{c|}{$\alpha = 8$}\\
			\cline{2-7}
			in \lineref{damp2} of \algref{deepECpr}  & PSNR & SSIM  & PSNR & SSIM & PSNR & SSIM \\
			\hline
			deterministic damping \eqref{damp}  & 38.05  & 0.9565 & 36.3 &  0.9468 & 34.20  & 0.9267 \\ 
			stochastic damping \eqref{sdamp} & {39.75}   & {0.9720}  & {37.01} &  {0.9567} & 34.86 & 0.9404 \\ 			\hline
	\end{tabular}}
	\label{tab:FFHQ_OSF_damping_ablation}
\end{table}

%

\begin{figure}[t]
	\centering
	\includegraphics[width=0.9\linewidth,trim=7 9 5 6,clip]{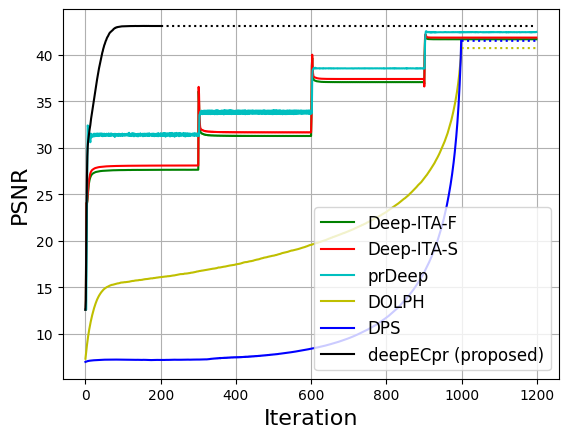}
	\caption{\textb{Average PSNR versus iteration for FFHQ phase retrieval from CDP measurements at noise level $\alpha=5$.}}
	\label{fig:PSNRvsIter_CDP_FFHQ}
\end{figure}

\begin{table*}[t]
	\centering
	\caption{Average PSNR and SSIM for natural and unnatural grayscale image phase retrieval with CDP measurements at noise level $\alpha$.}
	\resizebox{\linewidth}{!}{
		\begin{tabular}{|c|c|c|c|c|c|c|c|c|c|c|c|c|}
			\hline
			\multirow{3}{*}{method }& \multicolumn{6}{c|}{natural} & \multicolumn{6}{c|}{unnatural}\\
			\cline{2-13}
			& \multicolumn{2}{c|}{$\alpha = 5$} & \multicolumn{2}{c|}{$\alpha = 15$} & \multicolumn{2}{c|}{$\alpha = 45$}& \multicolumn{2}{c|}{$\alpha = 5$} & \multicolumn{2}{c|}{$\alpha = 15$} & \multicolumn{2}{c|}{$\alpha = 45$}\\
			\cline{2-13}
			& PSNR & SSIM  & PSNR & SSIM & PSNR & SSIM & PSNR & SSIM  & PSNR & SSIM & PSNR & SSIM \\
			\hline
			HIO  &  36.39 & 0.9541 & 26.65 & 0.7663 &  17.13 & 0.4155 & 36.45 & 0.8891 & 26.94 & 0.6133 & 18.55 & 0.3175\\ 
			Deep-ITA-F  & 38.67 & 0.9783 & 29.57  & 0.8595 & 22.77 & 0.6301 & 39.42 & 0.9607 & 29.80 & 0.7526 & 21.40 & 0.4352\\
			Deep-ITA-S  & {38.80} &  \underline{0.9797}  & 28.56 & 0.8357 &  17.25 & 0.4260  & {39.80} & {0.9679}  & 28.56 & 0.7077 & 18.74 &  0.3211 \\
			prDeep  & 38.73 & 0.9785 & \underline{32.49} & \underline{0.9388} & {26.38} &  \underline{0.8167} & 39.63 & 0.9660 &  \underline{33.84} & \underline{0.9267} & \underline{27.57} &  \underline{0.8220} \\
			\textb{DOLPH}  & \textb{\underline{39.19}} & \textb{0.9738} & \textb{28.37} & \textb{0.7435} & \textb{17.63} & \textb{0.3336} & \textb{\underline{40.40}} & \textb{\underline{0.9708}} &  \textb{29.11} & \textb{0.6571} & \textb{19.28} &  \textb{0.2797} \\
			\textb{DPS}  & \textb{37.66} & \textb{0.9617} & \textb{32.07} & \textb{0.9038} & \textb{\underline{26.76}} & \textb{0.7634} & \textb{39.05} & \textb{0.9603} &  \textb{33.15} & \textb{0.8978} & \textb{27.22} &  \textb{0.7423} \\
			deepECpr (proposed) & \textbf{39.46} & \textbf{0.9845}  & \textbf{32.75} &  \textbf{0.9439} & \textbf{27.06} & \textbf{0.8428} & \textbf{40.92} & \textbf{0.9808} &  \textbf{34.17} & \textbf{0.9356} & \textbf{28.14} &  \textbf{0.8292}\\
			\hline
	\end{tabular}}
	\label{tab:natural_unnatural_CDP}
\end{table*}

\begin{figure*}[t]
	\centering
	\includegraphics[width=\linewidth,trim=5 5 5 5,clip]{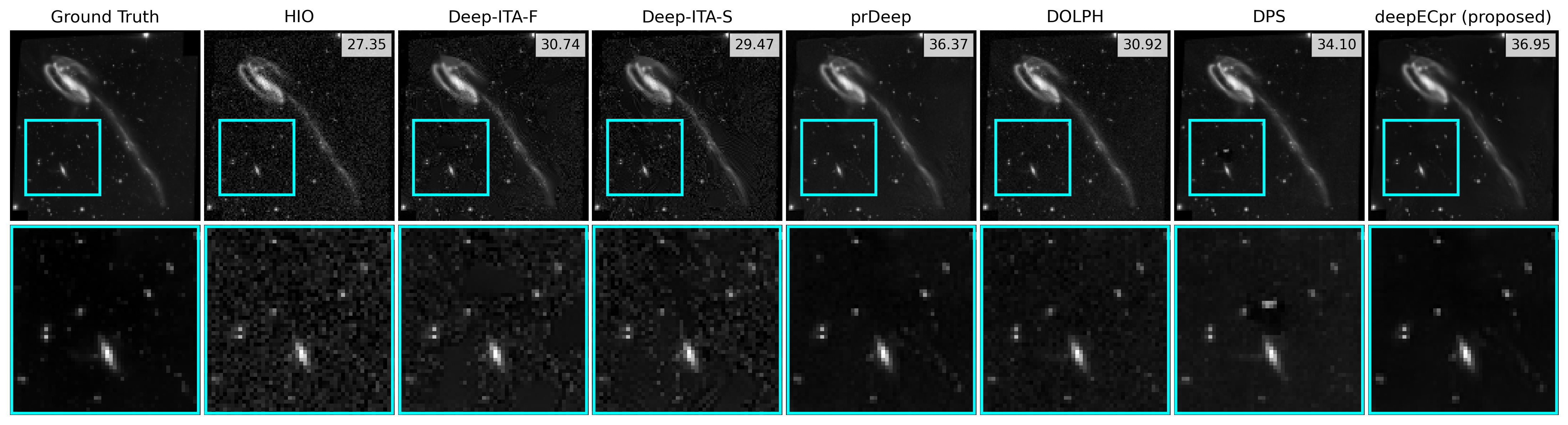}	
	\vspace{-5mm}
	\caption{\textb{Top: Example unnatural image recoveries from phaseless CDP measurements at noise level $\alpha=15$, with PSNR indicated in the top right corner of each image. Bottom: Zoomed versions of the cyan regions in the top row. Note that the HIO, Deep-ITA, and DOLPH recoveries are visibly noisy, and that DPS has hallucinated a bright spot in the middle of the image.}}
	\label{fig:unnatural_CDP_recon}
\end{figure*}

We now investigate the recovery of FFHQ test images from noisy, phaseless CDP measurements.
Deep-ITA, prDeep, \textb{DOLPH}, DPS, and deepECpr are configured as described in \secref{FFHQ_OSF}, except that
the Deep-ITA uses $\lambda=0.01\sqrt{v}$, \textb{DOLPH uses stepsize $5\times 10^{-6}$,} and DPS uses scale parameter $0.05$, \textb{in all three} cases to maximize validation PSNR.
deepECpr uses the denoising factor $\beta=0.002$.

\Tabref{FFHQ_OSF_CDP} shows the average PSNR and SSIM values at noise levels $\alpha \in\{5, 15, 45\}$.
There we see that deepECpr outperforms the other methods in both PSNR and SSIM at all $\alpha$ and gives a PSNR advantage of $>2$~dB at $\alpha=45$.

\Figref{FFHQ_CDP_recon1} shows example recoveries at $\alpha = 45$.
There we see that \textb{HIO, prDeep, and DOLPH} fall prey to noise-like artifacts.
Meanwhile, \textb{Deep-ITA-F over-smooths the image, while Deep-ITA-S and DPS introduce hallucinations.}  
Finally, deepECpr does the best job of faithfully recovering the true image.

\Figref{PSNRvsIter_CDP_FFHQ} plots the PSNR (averaged over the FFHQ test images) versus iteration at $\alpha = 5$.
As before, deepECpr converges \textb{significantly} faster than its competitors.

\subsection{CDP Phase Retrieval of Grayscale Images} \label{sec:grayscale_CDP}

We now investigate the recovery of natural and unnatural grayscale test images from noisy, phaseless CDF measurements.
The algorithms under test are configured as described in \secref{algs}.
In addition, prDeep uses the author-recommended $\lambda=0.1\sqrt{v}$, while Deep-ITA uses $\lambda=0.25\sqrt{v}$ to maximize validation PSNR. 
For deepECpr, we use the same blind DnCNN denoisers as prDeep and Deep-ITA, with $\beta=0.045,0.035,0.025,0.020$ for $\dncnn{60}, \dncnn{40}, \dncnn{20}, \dncnn{10}$, respectively. 
\textb{DOLPH and DPS utilize the ImageNet diffusion denoiser \cite{Dhariwal:NIPS:21} with step-size of $5\times 10^{-6}$ and a scale parameter of $0.025$, respectively, to maximize validation PSNR.}

Table~\ref{tab:natural_unnatural_CDP} shows the average PSNR and SSIM values at noise levels $\alpha \in\{ 5, 15, 45\}$. 
There deepECpr achieves the highest PSNR and SSIM values at all tested noise levels for both natural and unnatural grayscale images. 

\Figref{unnatural_CDP_recon} shows example recoveries at $\alpha=15$.
There we see that deepECpr's recovery looks cleaner and sharper than those of the competing methods.

\begin{table*}[t]
	\centering
	\caption{Average PSNR and SSIM for natural and unnatural grayscale test images with OSF phase retrieval at noise level $\alpha$.}
	\resizebox{\linewidth}{!}{
		\begin{tabular}{|c|c|c|c|c|c|c|c|c|c|c|c|c|}
			\hline
			\multirow{3}{*}{method }& \multicolumn{6}{c|}{natural} & \multicolumn{6}{c|}{unnatural}\\
						\cline{2-13}
			& \multicolumn{2}{c|}{$\alpha = 4$} & \multicolumn{2}{c|}{$\alpha = 6$} & \multicolumn{2}{c|}{$\alpha = 8$}& \multicolumn{2}{c|}{$\alpha = 4$} & \multicolumn{2}{c|}{$\alpha = 6$} & \multicolumn{2}{c|}{$\alpha = 8$}\\
						\cline{2-13}
			& PSNR & SSIM  & PSNR & SSIM & PSNR & SSIM & PSNR & SSIM  & PSNR & SSIM & PSNR & SSIM \\
			\hline
			HIO   &  24.32 &  0.6932 & 22.75 &  0.6267 &  21.63 & 0.5702 &  26.66 & 0.6221  & 25.10 & 0.5653 & 24.04 & 0.5231 \\  
			Deep-ITA-F  & 31.19 & 0.8996 & 29.64 & 0.8642 & 28.35  & 0.8278 & 28.30 & 0.6904  & 26.76 & 0.6466 & 26.58 & \underline{0.6450} \\
			Deep-ITA-S  & 30.98 & 0.8933 &  29.71 & 0.8603 &  28.78 & 0.8343 & 28.20 &  0.6894 & \underline{26.96}  & 0.6472 & \underline{26.75} & 0.6390 \\
			prDeep  & \underline{35.71} & \underline{0.9632} &  \underline{32.56} & \underline{0.9242} & \underline{29.57} & \underline{0.8670} & \underline{30.22}  & \underline{0.7533}  & 26.66 & \underline{0.6630} & 25.96 & 0.6297  \\
			\textb{DOLPH}  & \textb{16.65} & \textb{0.3331} & \textb{16.63} & \textb{0.3234} & \textb{16.61} & \textb{0.3208} & \textb{19.25} & \textb{0.4226} &  \textb{19.27} & \textb{0.4025} & \textb{19.08} &  \textb{0.3833} \\
			\textb{DPS}  & \textb{14.65} & \textb{0.2177} & \textb{14.22} & \textb{0.2014} & \textb{14.19} & \textb{0.1851} & \textb{17.19} & \textb{0.3168} &  \textb{15.66} & \textb{0.2742} & \textb{15.75} &  \textb{0.2774} \\
			deepECpr (proposed) & \textbf{37.30} & \textbf{0.9766} & \textbf{34.06 }& \textbf{0.9546} & \textbf{31.85} & \textbf{0.9302 } & \textbf{30.98 }& \textbf{0.7675} & \textbf{27.45} & \textbf{0.6897} & \textbf{27.09} & \textbf{0.6780} \\
			\hline
	\end{tabular}}
	\label{tab:natural_unnatural_OSF}
\end{table*}

\begin{figure*}[t]
	\centering
	\includegraphics[width=\linewidth,trim=5 5 5 5,clip]{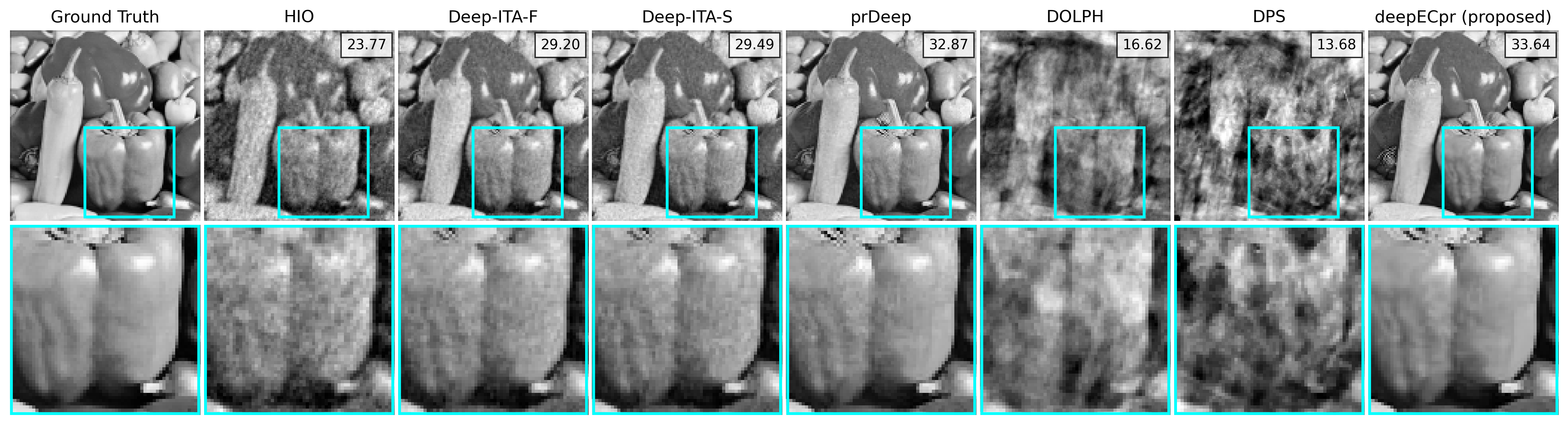}
	\vspace{-5mm}
	\caption{\textb{Top: Example natural image recoveries from phaseless OSF measurements at noise level $\alpha=8$, with PSNR indicated in the top right corner of each image. Bottom: Zoomed versions of the cyan regions in the top row.  Note that the HIO, Deep-ITA, DOLPH, and DPS recoveries contain strong artifacts, while the prDeep recovery shows an overly rough texture.}}
	\label{fig:natural_OSF_recon}
\end{figure*}

\subsection{OSF Phase Retrieval of Grayscale Images} \label{sec:grayscale_OSF}

Finally, we investigate the recovery of natural and unnatural grayscale test images from noisy, phaseless OSF measurements.
The algorithms are set up as described in \secref{grayscale_CDP}, except that prDeep and Deep-ITA use the author-recommended choices $\lambda=\sqrt{v}$ and $\lambda=0.025\sqrt{v}$, respectively, and deepECpr uses $\beta=0.45, 0.35, 0.25, 0.20$ for $\dncnn{60}, \dncnn{40}, \dncnn{20}, \dncnn{10}$ over $150,100,75,75$ iterations, respectively.
\textb{Additionally, DOLPH employs a step-size of $5\times 10^{-5}$, while DPS uses a scale parameter of $0.05$, each to maximize validation PSNR.}

\Tabref{natural_unnatural_OSF} shows the average PSNR and SSIM values at noise levels $\alpha \in \{4, 6, 8\}$. 
There, for both natural and unnatural test images, deepECpr achieves the highest PSNR and SSIM at all tested noise levels.
For natural images, deepECpr's PSNR exceeds the second-best PSNR by $\geq 1.5$~dB.

\Figref{natural_OSF_recon} shows example recoveries at $\alpha = 8$.
There we see that deepECpr recovers the fine details and texture of the original image, while the other approaches show noise-like artifacts and overly rough textures.


\section{Conclusion} \label{sec:conclusion}

In this paper, we proposed ``deepECpr,'' which combines expectation-consistent (EC) approximation with deep denoising networks to advance the state-of-the-art in phase retrieval. 
In addition to a non-traditional application of EC, deepECpr employs a novel stochastic damping scheme.
Experimental results with oversampled-Fourier and coded-diffraction-pattern operators, and with color, natural, and unnatural grayscale images, demonstrate deepECpr's advantages in image recovery from noise-corrupted phaseless measurements. 
In almost all cases, deepECpr yielded improved PSNR and SSIM over the state-of-the-art prDeep, Deep-ITA, \textb{DOLPH,} and DPS methods with \textb{significantly} fewer denoiser calls. 
Although not explicitly tested, the proposed deepEC approach should be directly applicable to other generalized-linear-model problems like dequantization, logistic regression, and image recovery in non-Gaussian noise.

\bibliographystyle{ieeetr}

\bibliography{macros_abbrev,stc,books,misc,comm,sparse,mri,machine,phase}

%

\appendices

\section{Variance of $p(z_i|y_i;\bar{z}\of{1}_i,\bar{v}\of{1})$}\label{app:post_var_z_of_1_i}

For brevity, we omit the variable index ``$i$" since the development is identical for all indices $i$, as well as the superscript $(\cdot)\of{1}$.
We set $\hat{v}$ at the approximation of the posterior variance given by the Laplace approximation \cite{Bishop:Book:07}, which is
\begin{align}
\tr\{\vec{H}(\hat{z})^{-1}\}
\text{~~for~~}
\vec{H}(\hat{z}) \defn 
\nabla_{z}^2(-\ln p(z|y;\bar{z},\bar{v}))\big|_{z = \hat{z} } ,
\end{align}
where $\hat{z}$ is the MAP estimate given in \eqref{MAP}, i.e.,
\begin{align}
\hat{z} 
= \frac{\bar{v}y + 2v|\bar{z}|}
       {\bar{v} + 2v}           e^{j \angle \bar{z}} ,
\end{align}
and $\nabla_{z}^2$ denotes the Hessian with respect to the real and imaginary parts of $z$. 

In~\appref{hess}, $\tr\{\vec{H}(\hat{z})^{-1}\}$ is derived to be
\begin{align}
\tr\big\{\vec{H}(\hat{z})^{-1}\big\}
&=v\frac{2\frac{|\hat{z}|}{y}-\frac{\bar{v}}{\bar{v}+2v}}
        {\frac{\bar{v}+2v}{\bar{v}}\frac{|\hat{z}|}{y} - 1}
\end{align}
Assuming that $y\geq -2v|\bar{z}|/\bar{v}$, we have 
\begin{align}
2\frac{|\hat{z}|}{y}-\frac{\bar{v}}{\bar{v}+2v}
&= \frac{2\bar{v} + 4v|\bar{z}|/y}{\bar{v} + 2v} - \frac{\bar{v}}{\bar{v}+2v} 
= \frac{\bar{v} + 4v|\bar{z}|/y}{\bar{v} + 2v}
\end{align}
and
\begin{align}
\frac{\bar{v}+2v}{\bar{v}}\frac{|\hat{z}|}{y} - 1
&= \frac{\bar{v}+2v}{\bar{v}}\cdot\frac{\bar{v} + 2v|\bar{z}|/y}{\bar{v} + 2v} - 1
= \frac{2v|\bar{z}|/y}{\bar{v}} 
\end{align}
so that
\begin{align}
\tr\big\{\vec{H}(\hat{z})^{-1}\big\}
&= v \frac{\bar{v} + 4v|\bar{z}|/y}{\bar{v} + 2v} \cdot \frac{\bar{v}}{2v|\bar{z}|/y} 
= \frac{\bar{v}(\bar{v}y + 4v|\bar{z}|)}{2|\bar{z}|(\bar{v} + 2v)} .
\end{align}

\section{Derivation of the Trace Inverse Hessian}\label{app:hess}

Here we derive $\nabla_{z}^2(-\ln p(z|y;\bar{z},\bar{v}))$, the Hessian of $-\ln p(z|y;\bar{z},\bar{v})$ with respect to the real and imaginary components of $z\in\Complex$.
We will use $z_r$ and $z_j$ to denote the real and imaginary components of $z$, and $\bar{z}_r$ and $\bar{z}_j$ to denote the real and imaginary components of $\bar{z}$, respectively. 
We have 
\begin{align}
\lefteqn{ -\ln p(z|y;\bar{z},\bar{v}) 
=  \tfrac{1}{2v}(y-|z|)^2 +  \tfrac{1}{\bar{v}}|{z}-\bar{z}|^2 + \text{const} }\\
&=  \tfrac{1}{2v}\Big(y-\sqrt{z_r^2+z_j^2}\Big)^2 +  \tfrac{1}{\bar{v}}\big[({z}_r - \bar{z}_r)^2  +  ({z}_j - \bar{z}_j)^2\big] 
\nonumber\\&\quad
+ \text{const} .
\end{align}
This implies that
\begin{eqnarray}
\lefteqn{ - \frac{\partial}{\partial z_r}\ln p(z|y;\bar{z},\bar{v}) }\nonumber\\
&=&\frac{1}{v}\Big(\sqrt{{z}_r^2 + {z}_j^2}-y\Big) \frac{1}{2} ({z}_r^2 + {z}_j^2)^{-1/2}2{z}_r 
+ \frac{2}{\bar{v}}({z}_r - \bar{z}_r) \qquad\\
&=&\frac{1}{v}\bigg(1 - \frac{y}{\ssqrt{{z}_r^2 + {z}_j^2}} \bigg){z}_r + \frac{2}{\bar{v}}({z}_r - \bar{z}_r),
\end{eqnarray}
and similarly that 
\begin{align}
	- \frac{\partial}{\partial z_j}
\ln p(z|y;\bar{z},\bar{v})
	&= \frac{1}{v}\bigg(1 - \frac{y}{\ssqrt{{z}_r^2 + {z}_j^2}} \bigg){z}_j + \frac{2}{\bar{v}}({z}_j - \bar{z}_j).
\end{align}
The second derivatives are
\begin{align}
\lefteqn{ -\frac{\partial^2}{\partial z_r^2} \ln p(z|y;\bar{z},\bar{v}) }\nonumber\\
&= \frac{\partial}{\partial z_r} \bigg( \frac{1}{v}\bigg(1 - \frac{y}{\ssqrt{{z}_r^2 + {z}_j^2}} \bigg){z}_r 
+ \frac{2}{\bar{v}}({z}_r - \bar{z}_r) \bigg)\\
&=  \frac{1}{v}\bigg(1 - \frac{y}{\ssqrt{{z}_r^2 + {z}_j^2}} +  \frac{yz_r}{2[{{z}_r^2 + {z}_j^2}]^{3/2}}2z_r \bigg) 
+ \frac{2}{\bar{v}}\\
&=  \frac{1}{v}\bigg(1 - \frac{y({z}_r^2 + {z}_j^2)}{[{{z}_r^2 + {z}_j^2}]^{3/2}} +  \frac{yz_r^2}{[{{z}_r^2 + {z}_j^2}]^{3/2}} \bigg) 
+ \frac{2}{\bar{v}}\\
&=  \frac{1}{v}\bigg(1 - \frac{y{z}_j^2}{|z|^{3}} \bigg) + \frac{2}{\bar{v}},
\end{align}
and 
\begin{align}
-\frac{\partial^2}{\partial z_j^2}\ln p(z|y;\bar{z},\bar{v})
&=  \frac{1}{v}\bigg(1 - \frac{y{z}_r^2}{|z|^{3}} \bigg) + \frac{2}{\bar{v}},
\end{align}
and 
\begin{align}
\lefteqn{ -\frac{\partial^2}{\partial z_j\partial z_r}\ln p(z|y;\bar{z},\bar{v}) }\nonumber\\
&= \frac{\partial}{\partial z_j} \bigg( \frac{1}{v}\bigg(1 - \frac{y}{\ssqrt{{z}_r^2 + {z}_j^2}} \bigg){z}_r 
+ \frac{2}{\bar{v}}({z}_r - \bar{z}_r) \bigg)\\
&=  \frac{1}{v}\frac{yz_r}{2[{{z}_r^2 + {z}_j^2}]^{3/2}}2z_j 
=  \frac{yz_rz_j}{v|z|^{3}},
\end{align}
and so the Hessian matrix  is
\begin{align}
\vec{H}(z) 
&= \mat{ -\frac{\partial^2}{\partial z_r^2}\ln p(z|y;\bar{z},\bar{v})& 
	-\frac{\partial^2}{\partial z_j\partial z_r}\ln p(z|y;\bar{z},\bar{v}) \\
	-\frac{\partial^2}{\partial z_j\partial z_r}\ln p(z|y;\bar{z},\bar{v}) &
	-\frac{\partial^2}{\partial z_j^2}\ln p(z|y;\bar{z},\bar{v})} \nonumber \\
&= 	\mat{
	\frac{1}{v}\big(1 - \frac{y{z}_j^2}{|z|^{3}} \big) + \frac{2}{\bar{v}}& 
	\frac{yz_rz_j}{v|z|^{3}}\\
	\frac{yz_rz_j}{v|z|^{3}} &
	\frac{1}{v}\big(1 - \frac{y{z}_r^2}{|z|^{3}} \big) + \frac{2}{\bar{v}}} \\
&= \bigg(\frac{1}{v}+\frac{2}{\bar{v}}\bigg)\vec{I}_2
    - \frac{y}{v|z|^3}\mat{z_j\\-z_r}\mat{z_j & -z_r}
\label{eq:hessian} .
\end{align}
Using the matrix inversion lemma,
\begin{align}
&\tr\big\{ (a\vec{I}_2+b\vec{cc}\tran)^{-1} \big\}
= \tr\Big\{ \frac{1}{a}\Big( \vec{I}_2 - \frac{b \vec{cc}\tran}{a+b\vec{c}\tran\vec{c}} \Big) \Big\}  \\
&= \frac{1}{a}\Big( 2 - \frac{b\vec{c}\tran\vec{c}}{a+b\vec{c}\tran\vec{c}}\Big)
= \frac{2+a^{-1}b\vec{c}\tran\vec{c}}{a+b\vec{c}\tran\vec{c}}
.
\end{align}
Applying this to \eqref{hessian} using 
\begin{align}
a &= \frac{1}{v}+\frac{2}{\bar{v}}
=\frac{\bar{v}+2v}{\bar{v}v} \\
b\vec{c}\tran\vec{c}
&=-\frac{y}{v|z|^3}\mat{z_j&-z_r}\mat{z_j\\-z_r} 
= -\frac{y}{v|z|}
,
\end{align}
we get
\begin{align}
\tr\big\{\vec{H}(z)^{-1}\big\}
&=\frac{2-\frac{\bar{v}v}{\bar{v}+2v} \frac{y}{v|z|}}{\frac{\bar{v}+2v}{\bar{v}v} - \frac{y}{v|z|}}
=v\frac{2\frac{|z|}{y}-\frac{\bar{v}}{\bar{v}+2v}}{\frac{\bar{v}+2v}{\bar{v}}\frac{|z|}{y} - 1}
\label{eq:tr_inv_hessian} .
\end{align}

\end{document}